\documentclass[10pt,journal,compsoc]{IEEEtran}

\ifCLASSOPTIONcompsoc
  \usepackage[nocompress]{cite}
\else
  \usepackage{cite}
\fi

\usepackage[table,xcdraw]{xcolor}

\usepackage{graphicx}
\newtheorem{theorem}{Theorem}
\newtheorem{proof}{Proof}
\usepackage{tikz}
\usepackage{comment}
\usepackage{amsmath,amssymb} 
\usepackage[colorlinks,linkcolor=red]{hyperref}
\usepackage{multirow}
\usepackage{multirow}
\usepackage{algorithm}  
\usepackage{algorithmicx}  
\usepackage{algpseudocode}  
\usepackage{amsmath}  
\usepackage{array}

\begin{document}

\title{Mitigating Accuracy-Robustness Trade-off via Balanced Multi-Teacher Adversarial Distillation}

\author{Shiji~Zhao,
        Xizhe~Wang, 
        and~Xingxing~Wei$^*$~\IEEEmembership{Member,~IEEE}
\IEEEcompsocitemizethanks{\IEEEcompsocthanksitem Shiji Zhao,  Xizhe Wang, Xingxing Wei were at the Institute of Artificial Intelligence, Beihang University, No.37, Xueyuan Road, Haidian District, Beijing,
100191, P.R. China. (E-mail: \{zhaoshiji123,  xizhewang, xxwei\}@buaa.edu.cn)
\protect\\
\IEEEcompsocthanksitem Xingxing Wei is the corresponding author.}
}

\markboth{IEEE TRANSACTIONS ON PATTERN ANALYSIS AND MACHINE INTELLIGENCE}%
{Shell \MakeLowercase{\textit{et al.}}: Bare Demo of IEEEtran.cls for Computer Society Journals}

\IEEEtitleabstractindextext{
\begin{abstract}
Adversarial Training is a practical approach for improving the robustness of deep neural networks against adversarial attacks. Although bringing reliable robustness, the performance towards clean examples is negatively affected after Adversarial Training, which means a trade-off exists between accuracy and robustness. 
Recently, some studies have tried to use knowledge distillation methods in Adversarial Training, achieving competitive performance in improving the robustness but the accuracy for clean samples is still limited. 
In this paper, to mitigate the accuracy-robustness trade-off, we introduce the Balanced Multi-Teacher Adversarial Robustness Distillation (B-MTARD) to guide the model's Adversarial Training process by applying a strong clean teacher and a strong robust teacher to handle the clean examples and adversarial examples, respectively. 
During the optimization process, to ensure that different teachers show similar knowledge scales, we design the Entropy-Based Balance algorithm to adjust the teacher's temperature and keep the teachers' information entropy consistent. Besides, to ensure that the student has a relatively consistent learning speed from multiple teachers, we propose the Normalization Loss Balance algorithm to adjust the learning weights of different types of knowledge. A series of experiments conducted on three public datasets demonstrate that B-MTARD outperforms the state-of-the-art methods against various adversarial attacks. 
\end{abstract}

}

\maketitle

\IEEEdisplaynontitleabstractindextext

%
\IEEEpeerreviewmaketitle

\IEEEraisesectionheading{\section{Introduction}\label{sec:introduction}}

\IEEEPARstart{D}{eep}  Neural Networks (DNNs) have become powerful tools for solving complex real-world learning problems, such as image classification \cite{he2016deep}, face recognition \cite{wang2018cosface}, and natural language processing \cite{sarikaya2014application}. However, Szegedy et al. \cite{szegedy2013intriguing} demonstrate that DNNs are vulnerable to adversarial attacks with imperceptible adversarial perturbations on input, which causes wrong predictions of DNNs. 
To defend against adversarial attacks, Adversarial Training is proposed and has shown its effectiveness in obtaining adversarial robust DNNs \cite{wang2019improving,croce2020reliable,madry2017towards}. While improving the robustness of DNNs, a negative impact exists on the model's accuracy on clean samples.
{Some methods are proposed to alleviate the trade-off between accuracy and robustness from different perspectives, e.g., optimization \cite{zhang2019theoretically,wang2019improving,pang2022robustness,yang2020closer,zhang2020geometry} and extra data \cite{alayrac2019labels,carmon2019unlabeled,hendrycks2019using}. } However, this phenomenon still exists and needs to be further explored.

Recently, to further enhance the robustness of small DNNs, Knowledge Distillation \cite{hinton2015distilling} is applied as a powerful tool in Adversarial Training, which can transfer knowledge from strong robust models to the student model. It utilizes the teacher's predictions as label information to guide the student model {within the framework of Adversarial Training, including the optimization to generate the adversarial examples (Maximization process) and apply adversarial examples to train the student with the assistance of the teacher (Minimization process).} These methods are called adversarial robustness distillation (ARD) \cite{goldblum2020adversarially,zhu2021reliable,zi2021revisiting}. Compared with one-hot ground truth labels, the teacher's predicted labels can not only retain the correctness of the target label but also reflect richer knowledge information of the non-target label. Several studies attempt to explain why knowledge distillation is effective through various views, e.g., Re-weighting \cite{furlanello2018born}, Privileged Information \cite{lopez2015unifying}, and Label-smoothing \cite{yuan2020revisiting}. Recently, Li et al. \cite{li2022asymmetric} argue that correctness, smooth regularization, and class discriminability in teacher's predicted labels can make up the knowledge of the teacher's predicted distribution, and that appropriate category discriminability can improve the effectiveness of knowledge distillation. While achieving impressive robust performance, we are curious if adversarial robustness distillation \cite{goldblum2020adversarially,zhu2021reliable,zi2021revisiting} could address the trade-off between accuracy and robustness.

Generally speaking, if we want to improve both accuracy and robustness through knowledge distillation, the provided teacher model is supposed to perform well in both aspects. That is to say, the teacher model should equally provide correct guidance and rich knowledge information for clean and adversarial examples. However, due to the existing trade-off, the teacher model itself is difficult to achieve the above goals. Therefore, a reasonable approach is to decouple adversarial knowledge distillation, i.e.,  using two teachers that are good at accuracy and robustness respectively (they can be called the clean teacher and the robust teacher) to guide the student in different knowledge types. Under this divide-and-conquer strategy, the student model guided by multiple teacher models can enhance its performance in both aspects theoretically. 

However, the actual implementation of this idea is challenging. Owing to the intrinsic trade-off between accuracy and robustness, the clean teacher and robust teacher push the student model in opposite directions. This is obviously different from the current multi-teacher knowledge distillation methods \cite{son2021densely, yuan2021reinforced,liu2020adaptive}, where their optimization goal of different teachers is consistent, i.e., all teachers push the student towards the good clean accuracy. The trade-off in our task leads to a difficult optimization mainly in two aspects: 
\textcolor{red}{(1)} How to make two teacher models have similar knowledge scales. When one teacher has too much knowledge scale, the student will be forced to obtain more knowledge from this teacher and relatively less knowledge from another, which directly results in an imbalance between accuracy and robustness. However, finding a clean teacher and a robust teacher with similar knowledge scales is not easy because the different network architectures and training methods will lead to a natural gap in knowledge scales between different teachers. {Thus, we need a metric to accurately measure and balance teachers' knowledge scales.} 
\textcolor{red}{(2)} How to balance the student's learning ability from different types of teachers' knowledge. In our multi-teacher setting, because two opposite optimization directions exist in adversarial robustness distillation, the student may over-fit an ability that is easier to learn but ignores another due to different learning difficulties. Only the cooperation of the student and teachers can help the student obtain both strong accuracy and robustness.  {Thus, we need an adaptive balance mechanism to adjust the learning process.} 

Based on the above discussions, in this paper, we propose the \textbf{Balanced Multi-Teacher Adversarial Robustness Distillation (B-MTARD)} to enhance both accuracy and robustness at the same time. During Adversarial Training, a clean teacher is to instruct the student for handling clean examples and a strong robust teacher is to instruct the student for handling adversarial examples. To meet the two challenges above, the novel Entropy-Based Balance and Normalization Loss Balance strategies are presented {to achieve a balanced state between accuracy and robustness.}

For the first challenge, to ensure that the two teacher models show similar knowledge scales during the optimization, we propose an \textbf{Entropy-Based Balance} algorithm to balance the information entropy among different teachers' predicted distribution. Specifically, based on information theory \cite{shannon1948mathematical}, relative entropy can measure the incremental information from an initialized network's predicted distribution to the well-trained teachers' predicted distribution, so we utilize the relative entropy to measure the teachers' knowledge scales. Furthermore, we theoretically prove that the difference in knowledge scales can be transformed into the difference in teachers' information entropy.  Thus Entropy-Based Balance algorithm can balance the knowledge scales by adjusting each teacher's temperature until the information entropy is appropriate and consistent.

For the second challenge, to maintain the relative equality of the student's learning speed from multiple teachers, we propose a \textbf{Normalization Loss Balance} algorithm to control loss weight and balance the influence between the clean teacher and the robust teacher. Specifically, inspired by \cite{chen2018gradnorm}, we design a relative loss to measure the knowledge proportion that the student learns from different teachers. The different loss weights for the update of student parameters are dynamically adjusted to keep the relative loss consistent during the training process. Compared with the initial state, the student model will be forced to learn relatively equal knowledge scales from all the teacher models. Our code is available at \url{https://github.com/zhaoshiji123/MTARD-extension}.

The main contributions of this work are three-fold:
\begin{itemize}
\item We propose a novel framework: Balanced Multi-Teacher Adversarial Robustness Distillation (B-MTARD). B-MTARD decouples the adversarial knowledge distillation to obtain a balance between accuracy and robustness. We apply a clean teacher and a robust teacher to adaptively bring both clean and robust knowledge to the student.
\item We propose the Entropy-Based Balance algorithm and Normalization Loss Balance algorithm to balance multiple teachers. The Entropy-Based Balance algorithm is applied to control the teachers' knowledge scales by dynamically adjusting temperatures. The Normalization Loss Balance algorithm is utilized to ensure that the student has relatively equal learning speeds for different teachers' knowledge.
\item We empirically verify the effectiveness of B-MTARD in improving performance. The Weighted Robust Accuracy (a measure to evaluate both clean and robust accuracy) of our B-MTARD trained models improves significantly against a variety of attacks compared to state-of-the-art Adversarial Training and knowledge distillation methods. 
{Besides, we show B-MTARD can achieve an obvious improvement compared to MTARD proposed in our conference version.}
\end{itemize}

This journal paper is an extended version of our ECCV paper (MTARD) \cite{zhao2022enhanced}. Compared with the conference version, we have made significant improvements and extensions in this version in the following aspects: \textcolor{red}{(1)} At the idea level, we discuss the trade-off problem in a more comprehensive view from both teachers and the student, while the conference version only considers student's view. We point out that the difference in two teachers' knowledge scales will affect the trade-off between accuracy and robustness, and should formulate this factor into our method (the first challenge in Section \ref{sec:introduction}). \textcolor{red}{(2)} {At the method level, a multiple teacher's knowledge adjustment mechanism named the Entropy-Based Balance algorithm is designed, where we first give the definition and metric for the knowledge scale and then present the balancing method (Section \ref{subsec:Entropy-Based Balance in MTARD}). }\textcolor{red}{(3)} {At the experimental level,  we give a comprehensive comparison between B-MTARD with our conference version MTARD to show the advantage}, and also additionally conduct experiments on the Tiny-ImageNet dataset and compare our B-MTARD with more SOTA methods against more advanced attacks. In addition, we give more ablation studies to comprehensively test our method (Section \ref{sec:Experiment}). 

The rest of the paper is organized as follows: Related work is given in Section \ref{sec:Related Work}. Section \ref{sec:Methodology} introduces the details of our B-MTARD. The experiments are conducted in Section \ref{sec:Experiment}, and the conclusion is given in Section \ref{sec:Conclusion}.

\section{Related Work}
\label{sec:Related Work}
\subsection{Adversarial Attacks}

Since Szegedy et al. \cite{szegedy2013intriguing} propose that adversarial examples can mislead the deep neural network, lots of effective adversarial attack methods, such as the Fast Gradient Sign Method (FGSM) \cite{goodfellow2014explaining}, Projected Gradient Descent Attack (PGD) \cite{madry2017towards}, and Carlini and Wagner Attack (CW) \cite{carlini2017towards} are proposed. Attack methods can be divided into white-box attacks and black-box attacks. White-box attacks know all the parameter information of the target model when generating adversarial examples, and black-box attacks know little or no information. 
In general, black-box attacks simulate the model gradient by repeatedly querying the target model (query-based attack) \cite{andriushchenko2020square,wei2022sparse,wei2022adversarial,wei2022simultaneously,xingxing2023efficient,wei2021black,liang2022parallel} or searching for an alternative model similar to the target model (transfer-based attack) \cite{demontis2019adversarial,huang2019enhancing}.
Recently, a strong adversarial attack method named AutoAttack (AA) \cite{croce2020reliable} is proposed, which consists of four attack methods, including Auto-PGD (APGD), Difference of Logits Ratio (DLR) attack, FAB-Attack \cite{croce2020minimally}, and the black-box Square Attack \cite{andriushchenko2020square}.

\subsection{Adversarial Training}

Adversarial Training \cite{madry2017towards,zhang2019theoretically} is seen as an effective way to defend against adversarial attacks. Madry et al. \cite{madry2017towards} formulate Adversarial Training as a min-max optimization problem as follows:
\begin{align}
\label{eq:1}
\min\limits_{\theta}E_{(x,y)\sim \mathcal{D}}[\max\limits_{\delta\in\Omega}\mathcal{L} (f(x+\delta;\theta),y)],
\end{align}
where $f$ represents a deep neural network with weight $\theta$, $D$ represents a distribution of the clean example $x$ and the ground truth label $y$.  $\mathcal{L}$
represents the loss function. $\delta$ represents the adversarial perturbation, and $\Omega$ represents a bound, which can be defined as $\Omega = \left\{\delta: ||\delta||\leq \epsilon \right\}$ with the maximum perturbation scale $\epsilon$.

{To alleviate the trade-off between accuracy and robustness, some methods are proposed from different perspectives, e.g., optimization  \cite{zhang2019theoretically,wang2019improving,pang2022robustness,yang2020closer} and extra data \cite{alayrac2019labels,carmon2019unlabeled,hendrycks2019using}.  From the perspective of optimization,
Zhang et al. \cite{zhang2019theoretically} attempt to reduce the gap between accuracy and robustness by minimizing a Kullback–Leibler (KL) divergence loss (TRADES). Wang et al. \cite{wang2019improving} further improve performance through Misclassification-Aware Adversarial Training. Stutz et al. \cite{stutz2019disentangling} claim that manifold analysis can be helpful in achieving accuracy and robustness. Yang et al. \cite{yang2020closer} argue that the trade-off can be mitigated by optimizing the local Lipschitz functions. Pang et al. \cite{pang2022robustness} propose to employ local equivariance to describe the ideal behavior of a robust model, which facilitates the reconciliation between accuracy and robustness (SCORE). From the perspective of data, \cite{alayrac2019labels} and \cite{carmon2019unlabeled} find that additional unlabeled data can be useful to improve both the accuracy and robustness. \cite{hendrycks2019using} shows that adversarial pre-training with extra data can remarkably improve accuracy and adversarial robustness.
}

{Different from the previous methods, we try to alleviate the trade-off by utilizing multi-teacher adversarial distillation, which applies a clean teacher and a robust teacher to bring both clean and robust knowledge to the student.}


\subsection{Knowledge Distillation}
Knowledge distillation can transfer the performance of other models to the target model \cite{hinton2015distilling}. Extensive research has been widely studied in recent years \cite{kwon2020adaptive,yuan2021reinforced,li2022asymmetric}. 
Knowledge distillation can briefly be formulated as the following optimization:
\begin{align}
\label{eq:2}
\mathop{\arg\min}_{\theta_S}(1-\alpha)CE (S(x),y)+\alpha\tau^2KL(S(x;\tau),T(x;\tau)),
\end{align}
where $S$ represents the student model with weight $\theta_S$, $T$ represents the teacher model. $KL$ is Kullback–Leibler divergence loss,  $CE$ represents the cross-entropy loss. $\tau$ is a temperature constant combined with softmax operation, $\alpha$ is a weight hyper-parameter.

{The original knowledge distillation \cite{hinton2015distilling} attempts to improve the student's performance on clean examples, and pre-define a high temperature until the teacher produces suitably soft labels, while the same high temperature is applied to train the student to match these soft labels, which allows the student to better acquire knowledge from a single type of teacher. Our method (with our Entropy-Based
Balance Algorithm) is to balance the student's performance between accuracy and robustness, and we adaptively adjust the temperature for different types of teachers to balance their knowledge scales, which allows the student to equally acquire knowledge from different types of teachers.}

The temperature $\tau$ is artificially adjusted as a hyper-parameter in previous work. Recently, \cite{li2022curriculum,liu2022meta} propose to adjust $\tau$ automatically based on the designed learnable sub-network. These methods need to train the sub-network according to the feedback of the student model, which adds additional training overhead. Our method automatically adjusts the temperature only based on multiple teachers' knowledge scales and has almost no computational cost. 

Some studies also exist on multi-teacher knowledge distillation, and are designed from different views: including response-based knowledge \cite{kwon2020adaptive,yuan2021reinforced}, feature-based knowledge \cite{liu2020adaptive,zhao2020highlight}, and relation-based knowledge \cite{you2017learning,wu2019distilled}.
Previous methods often use multiple teachers for knowledge distillation for the same examples, and those teachers tend to optimize the student in similar directions. Different from previous methods, our multi-teacher adversarial robustness distillation targets two different kinds of examples. Since adversarial examples are generated to mislead the model, there is a huge gap between the optimization direction of clean examples and adversarial examples, which exists great difficulties compared with previous research.

\begin{figure*}[ht]
  \centering
\includegraphics[width=0.95\textwidth]{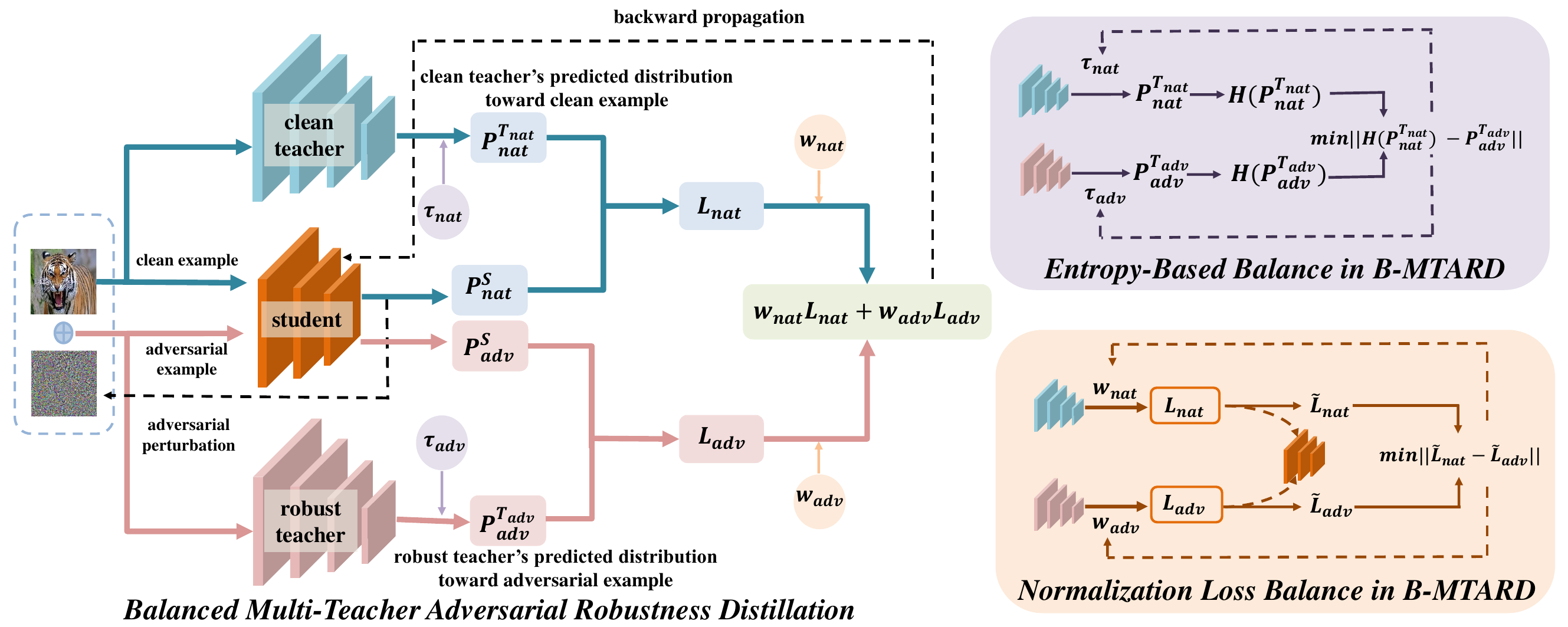}\\
\caption{The framework of our Balanced Multi-Teacher Adversarial Robustness Distillation (B-MTARD). In the training process, we first generate adversarial examples from the student. Then, with the guidance of the clean teacher and the robust teacher, the student is trained on clean examples and adversarial examples, respectively. Here, we apply the Entropy-Based Balance algorithm to adjust the teacher's knowledge scales; In addition, we use the Normalization Loss Balance algorithm to balance the student's knowledge learning speed from different teachers.}
\label{fig:framework}
\end{figure*}

\subsection{Adversarial Robustness Distillation}
To defend against the adversarial attack, defense distillation \cite{papernot2016distillation} is proposed to apply the traditional knowledge distillation. However, this method is considered as ``Obfuscated Gradients'' \cite{athalye2018obfuscated} and can be broken by CW attack \cite{carlini2017towards}.
Then a series of methods \cite{zi2021revisiting,goldblum2020adversarially,zhu2021reliable} apply knowledge distillation to further enhance the robustness based on Adversarial Training \cite{madry2017towards}. 
ARD \cite{goldblum2020adversarially} first proposes that using a strong robust model as the teacher model in the framework of Adversarial Training can achieve pretty robustness. IAD \cite{zhu2021reliable} performs adversarial knowledge distillation by compositing with unreliable teacher guidance and student introspection. RSLAD \cite{zi2021revisiting} uses the soft labels generated by the robust teacher model to produce adversarial examples, and further utilizes the robust labels to guide the training process of both clean and adversarial examples, which can effectively improve the robustness of the student. {Fair-ARD \cite{Yue2023Revisiting} is proposed to explore the fairness issues in ARD and enhance robust fairness by re-weighting different classes. Inspired by this paper, ABSLD \cite{zhao2023improving} tries to obtain a model with robust fairness by reducing the student's class-wise error risk gap and adjusting the class-wise smoothness degree of teacher labels via re-temperating different classes. }

{The above methods focus on improving the robustness or fairness of ARD. In our method, we consider both accuracy and robustness through clean and robust teachers to further mitigate the trade-off, which exists an obvious difference compared with other methods.}

\section{Methodology}
\label{sec:Methodology}
In this section, we propose B-MTARD to enhance both accuracy and robustness by applying the clean teacher and the robust teacher to guide the student. From the teachers' teaching view, the Entropy-Based Balance algorithm is proposed to balance the teacher's knowledge scale. From the student's learning view, the Normalization Loss Balance algorithm is designed to balance the student's learning speeds from different teachers.  


\subsection{Balanced Multi-Teacher Adversarial Distillation}

The training process in B-MTARD can be considered as a min-max optimization based on Adversarial Training. The whole framework can be viewed in Fig. \ref{fig:framework}. In the inner maximization, the adversarial examples are generated by the student based on the clean examples. In the outer minimization, 
the clean teacher and the robust teacher generate predicted labels to guide the student training process toward the clean examples and the adversarial examples, respectively. The optimization of the basic B-MTARD is defined as follows:
\begin{align}
\label{eq:4}
\begin{split}
    \mathop{\arg\min}_{\theta_S}(1-\alpha)&KL (S(x_{nat};\tau_{s}),T_{nat}(x_{nat};\tau_{nat}))  
    \\&+\alpha KL(S(x_{adv};\tau_{s}),T_{adv}(x_{adv};\tau_{adv})),
\end{split}
\end{align}
\begin{align}
\label{eq:5}
x_{adv}=\mathop{\arg\max}_{\delta\in\Omega}CE (S(x_{nat}+\delta),y),
\end{align}
where $x_{nat}$ and $x_{adv}$ are clean examples and adversarial examples, $S(x;\tau_{s})$ represents student network $S$ with temperature $\tau_{s}$. $T_{nat}(x;\tau_{nat})$ and $T_{adv}(x;\tau_{adv})$ represent the clean and robust teacher with temperature $\tau_{nat}$ and $\tau_{adv}$, respectively. $\alpha$ is a constant in the basic proposal. $CE$ denotes the Cross-Entropy loss in the maximization. Here, $\tau_{s}$, $\tau_{nat}$, and $\tau_{adv}$ are usually pre-defined in previous works.

The goal of B-MTARD is to obtain a student with strong accuracy as the clean teacher and strong robustness as the robust teacher. In the actual operation process, however, two teachers may display different knowledge scales; Meanwhile, the student may have different learning speeds toward different types of teachers' knowledge. The above two points may lead to an unbalanced performance in accuracy and robustness. So how to handle multi-teacher adversarial robustness distillation becomes a problem to be solved in the following two subsections.

\subsection{Entropy-Based Balance in B-MTARD}
\label{subsec:Entropy-Based Balance in MTARD}
\subsubsection{Measuring Knowledge Scale} \label{meansuring knowledge scale}

Due to the different model architectures and training methods, the teachers may have different knowledge scales. Eliminating the accuracy-robustness trade-off will become unrealistic if the student learns from the two teachers with no equal knowledge scales. 
Then we want to know how to measure the knowledge scale and further balance it.

{ In the knowledge distillation process, the part that teachers applied to guide students is the predicted distribution for the samples. So we can reasonably believe that the teacher’s knowledge exists in the predicted distribution for the samples, which actually represents the information understood by the teacher, e.g., the samples' classes, the samples' difficulty, or the relationship between the classes.
In knowledge distillation framework \cite{hinton2015distilling}, the teacher's predicted distribution can be defined as $P=\{p_{1}(x),..., p_{C}(x)\}$, where the predicted possibility $p_k(x)$ of $k$-th class can be formulated as follows:}
\begin{align}
\label{eq:3}
p_k(x) = \frac{exp(z_k(x)/\tau)}{\sum_{j=1}^Cexp(z_j(x)/\tau)},
\end{align}
{where $z_k(x)$ denotes the $k$-th dimension output logits of the model before the softmax layer, $\tau$ denotes the temperature applied in the training process of knowledge distillation.}

\begin{figure}[t]
  \centering
\includegraphics[width=0.49\textwidth]{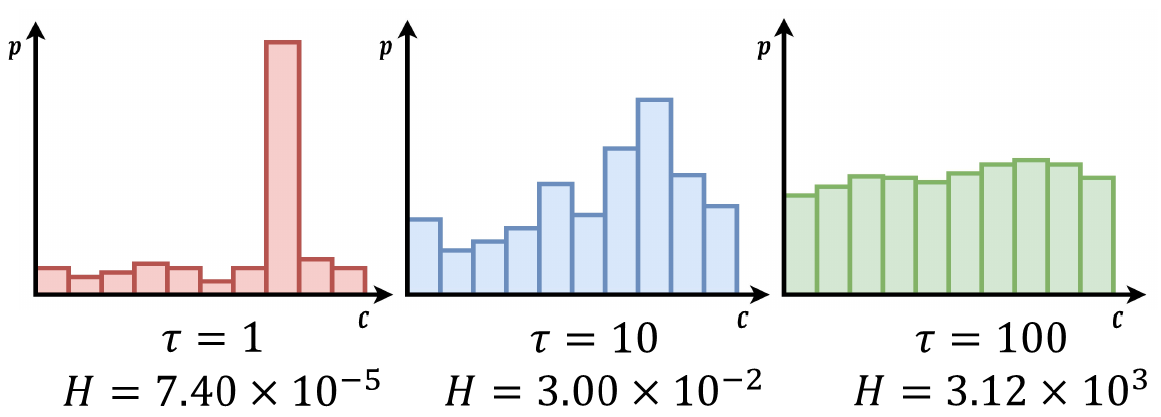}\\
\caption{Schematic diagram of the same logits but dealing with different temperatures $\tau$ and getting different predicted probabilities $p$. From the figure, the temperature $\tau$ can impact the model prediction and can further impact the information entropy $H$.}
\label{image:temp_entropy}
\end{figure}

{Although the knowledge described above exists in the teacher's predicted distributions, how to quantify them is not easy.  Here we try to consider it from the optimization perspective. Intuitively speaking, a well-trained network model is trained from a network model with randomly initialized weights. The model with randomly initialized weights can be considered as a general model $I$ without any knowledge; After training, the model can be considered to possess the knowledge brought by the optimization processes. Based on information theory, relative entropy (also known as Kullback–Leibler divergence \cite{shannon1948mathematical}) represents the information cost required from one distribution to another distribution. So the obtained knowledge scale can be denoted as the relative entropy from the predicted distribution of an initialized model $I$ to a well-trained model. Here we further define the teacher's knowledge scale $K^{T}$ in mathematical form as follows}:
\begin{align}
\label{knowledge defination}
K^{T}= KL(I(x),T(x)), 
\end{align}
{then we try to further simplify the knowledge scale $K^{T}$ and provide Theorem \ref{throrem}. And the corresponding proof can be found in the Appendix A (Proof A.1.).}
{
\begin{theorem}
\label{throrem}
The teacher’s knowledge scale $K^{T}$ is negatively related to the information entropy $H(P^{T})$ of teacher’s predicted distribution and have a relationship as follows:
\begin{align}
\label{knowledge with entropy}
K^{T}=logC-H(P^{T}), 
\end{align}
where$P^{T}=\{p_{1}^{T}(x),..., p_{C}^{T}(x)\}$ denote the teacher model $T$ predicted distribution, and $logC$ is a constant.
\end{theorem}
}
{Theorem \ref{throrem} implies that we can directly use information entropy to measure the teacher's knowledge scales. From the perspective of information theory, information entropy is used to describe the degree of randomness and uncertainty of information, so it is reasonable to apply information entropy for quantifying the knowledge scale.}

{Then we further analyze the factors that affect information entropy. Based on Eq.(\ref{eq:3}), the information entropy of predicted distribution is affected by two parts, one is the output logits $z(x)$, and the other is the temperature $\tau$, both of them can change the information entropy to influence the knowledge scale. The output logits $z(x)$ is determined by the model itself, when the teacher is selected, $z(x)$ cannot be changed.  While the temperature can be used as a hyper-parameter to directly change information entropy.}

In order to further explore the relationship between information entropy and temperature, we calculate the partial derivative of information entropy $H(P)$ with respect to temperature $\tau$ and can get the results as follows:
\begin{align}
\label{eq:15}
\nabla_{\tau}H(P) = \frac{(\sum_{j=1}^Cq_{j})(\sum_{j=1}^Cq_{j}log^2q_{j})-(\sum_{j=1}^Cq_{j}logq_{j})^2}{\tau(\sum_{j=1}^Cq_{j})^2},
\end{align}
where the $q$ represents the $exp(z(x)/\tau)$ in Eq.(\ref{eq:3}), and the more detailed derivation can be found in Appendix A (Proof A.2.). Meanwhile, we prove that the partial derivative $\nabla_{\tau}H(P) \geq 0$, and this conclusion is proved in Appendix A (Proof A.3.), thus the information entropy $H(P)$ shows a monotonically increasing relationship with temperature $\tau$. The effect is as shown in Fig. \ref{image:temp_entropy}: when the temperature increases, the prediction's information entropy also increases obviously.
{Then combined with Theorem \ref{throrem}, a larger temperature will lead to a smaller knowledge scale.}

\subsubsection{Balancing Knowledge Scale}
{After measuring the knowledge scale, we want to know the knowledge scales of different types of teachers, we conduct quantitative statistics, and the results are shown in Table \ref{table:teacher's knowledge}. The results denote that an obvious difference exists in the knowledge scale between the clean teacher and the robust teacher if without additional adjustment.
So it is very necessary to balance the knowledge scale between different teachers.}
Only when two teachers appear to be equally knowledgeable, the student will equally learn abilities from both the clean teacher and the robust teacher, and finally mitigate the trade-off between accuracy and robustness. 
\begin{table}[t] 
\begin{center}
\caption{{The quantitative statistic of knowledge scale for clean teacher ResNet-56 and robust teacher WideResNet-34-10 on CIFAR-10. We count the teachers’ knowledge scales before and after the adjustment of the Entropy-Based balance algorithm. The results are based on the statistics of the teachers’ knowledge scales in the final training epoch.
}}
\label{table:teacher's knowledge}
\begin{tabular}{m{2.4cm}<{\centering}m{2.6cm}<{\centering}m{2.6cm}<{\centering}}
\hline
  Teacher & Knowledge Scale (No Balance) & Knowledge Scale (After Balance)   \\
\hline
 ResNet-56 & 2.2518 & 1.2480  \\
WideResNet-34-10 & 1.3379 & 1.3249   \\
\hline
\end{tabular}
\end{center}
\end{table}

Based on the above analysis, we can balance the clean teacher's knowledge scale and robust teacher's knowledge scale by minimizing the gap of teachers' information entropy. Here we define the minimization of  knowledge scales' gap $\Delta_{K}$ as optimization objective in mathematical form:
\begin{align}
\label{eq:12}
\min\Delta_{K} = ||H(P_{adv}^{T_{adv}})- H(P_{nat}^{T_{nat}})||,
\end{align}
where $P_{nat}^{T_{nat}}$ and $P_{adv}^{T_{adv}}$ denote the clean teacher's predicted distribution towards $x_{nat}$ and adversarial teacher's predicted distribution towards $x_{adv}$, respectively.

Based on the above analysis in Section \ref{meansuring knowledge scale}, adjusting temperature can directly transform the information entropy of the model's prediction with negligible calculation cost, while adjusting logits $z(x)$ needs an additional calculation consumption for re-train models. So we select to solve the above optimization goal Eq. (\ref{eq:12}) by updating the multiple teachers' temperatures.



Then we need to apply the gradient of the expected object $\Delta_{K}$ concerning the clean teacher's temperature $\tau_{nat}$ and the robust teacher's temperature $\tau_{adv}$ and update them according to the gradient descent principle in the training process. Here we apply the optimization process of temperature $\tau_{nat}$ to expand our method (the optimization process of temperature $\tau_{adv}$ is similar, and we do not repeat the description), which is formulated as follows:
\begin{align}
\label{eq:13}
\tau_{nat} = \tau_{nat} - r_\tau \nabla_{\tau_{nat}} \Delta_{K},
\end{align}
we use the learning rate $r_\tau$ to control the magnitude of the gradient change, and then we further expand the preliminary derivation:
\begin{align}
\label{eq:14}
\begin{split}
\nabla_{\tau_{nat}} \Delta_{K}= \left \{
\begin{array}{ll}
    \nabla_{\tau_{nat}}H(P_{nat}^{T_{nat}}),             &H(P_{nat}^{T_{nat}}) \geq H(P_{adv}^{T_{adv}}),\\
    -\nabla_{\tau_{nat}}H(P_{nat}^{T_{nat}}),             &H(P_{nat}^{T_{nat}}) < H(P_{adv}^{T_{adv}}).
\end{array}
\right.
\end{split}
\end{align}

As described before (Eq.(\ref{eq:15})),  the information entropy $H(P)$ shows a monotonically increasing relationship with the temperature $\tau$.
So based on the consideration of the stable update, we can reasonably use the constant 1 to estimate both the $\nabla_{\tau_{nat}}H(P_{nat}^{T_{nat}})$ and $\nabla_{\tau_{adv}}H(P_{adv}^{T_{adv}})$ in the training process of updating the teacher's temperature, and we use $r_\tau$ to control the update range. 

Then combined the above proof with Eq. (\ref{eq:13}) and Eq. (\ref{eq:14}), the formula for updating the $\tau_{nat}$ is as follows:
\begin{align}
\label{eq:16}
\tau_{nat} = \tau_{nat} - r_\tau sign(H(P_{nat}^{T_{nat}}) - H(P_{adv}^{T_{adv}})),
\end{align}
where $sign(.)$ denotes the sign function. According to the same derivation process, we can also obtain the formula for updating the temperature $\tau_{adv}$ for teacher $T_{adv}$ can be formulated as follows:
\begin{align}
\label{eq:17}
\tau_{adv} = \tau_{adv} - r_\tau  sign(H(P_{adv}^{T_{adv}}) - H(P_{nat}^{T_{nat}})).
\end{align}

On a practical level, our Entropy-Based Balance algorithm can balance the teacher's knowledge scale by controlling the information entropy. {When the information entropy of teachers is not equal, both teachers' temperatures will be updated until the difference is eliminated: if a teacher has a high information entropy, then the teacher's temperature will reduce to further reduce the information entropy, leading to the higher knowledge scale based on Eq.(\ref{knowledge with entropy}); on the contrary, if a teacher has a low information entropy, then the teacher's temperature will increase to further increase the information entropy, leading to the lower knowledge scale.}
With the Entropy-Based Balance, the clean teacher and the robust teacher can appear equally knowledgeable when imparting knowledge to the student, which is helpful to mitigate the accuracy-robustness trade-off.

It should be mentioned that we only update the teachers' temperatures ($\tau_{nat}$ and $\tau_{adv}$) and keep the student's temperature ($\tau_{s}$) unchanged in the Entropy-Based Balance. In addition, in order to prevent information from being over-regulated, we set an upper and lower bound for the temperature to prevent the following situations: If the temperature is too small, the teacher's predicted distribution will be infinitely close to the one-hot label's distribution; When the temperature is too large, the teacher's predicted distribution will be close to the uniform distribution.

\subsection{Normalization Loss Balance in B-MTARD}

Although the two teachers are equally knowledgeable after the adjustment of the Entropy-Based Balance algorithm, the student may not learn equal knowledge from the two teachers during the training process, since the difficulties of recognizing clean examples and adversarial examples for the student are different: Adversarial examples are generated to mislead the student based on clean examples. In order to let the student get equal knowledge from the clean teacher and the robust teacher, a strategy is needed to control the student's relative learning speed toward different types of knowledge.
Inspired by gradient regularization methods in multi-task learning \cite{chen2018gradnorm}, we propose an algorithm to control the relative speed by dynamically adjusting the loss weights in the entire training process, which is called the Normalization Loss Balance algorithm.

On the mathematical level, the total loss in B-MTARD ultimately used for the student update at time $t$ can be represented as $L_{total}(t)$, which can be formulated as follows:
\begin{align}
\label{eq:18}
L_{total}(t) = w_{nat}(t) L_{nat}(t)+w_{adv}(t) L_{adv}(t), 
\end{align}
where the $L_{nat}(t)$ and $L_{adv}(t)$ denote the KL loss mentioned in Eq. (\ref{eq:4}), and the $w_{nat}(t)$ and $ w_{adv}(t)$ denote the constant hyper-parameters $1-\alpha$ and $\alpha$ mentioned in Eq. (\ref{eq:4}) in our task without the Normalization Loss Balance. 
Similar to the analysis in section \ref{subsec:Entropy-Based Balance in MTARD}, the knowledge scales' gap between the student and the teacher can be directly reflected by the value of the clean loss $L_{nat}(t)$ and adversarial loss $L_{adv}(t)$, and the vital to control the student's knowledge from multiple teachers consistent is controlling the loss weight of $w_{nat}(t)$ and $w_{adv}(t)$, which directly influence the student's learning of $L_{nat}(t)$ and $L_{adv}(t)$.  

Then our goal is to place $L_{nat}(t)$ and $L_{adv}(t)$ on a common metric through their relative magnitudes. Then based on the common metric, we can identify specific optimization directions to ensure that $L_{nat}(t)$ and $L_{adv}(t)$ have a relatively fair descent after the entire update process, and the final trained model can be equally affected by different parts of the total loss. 

To select the criterion metric to measure the decline of multiple losses, we choose a relative loss $\tilde{L}(t)$ following Chen et al. \cite{chen2018gradnorm}, which is defined as follows:
\begin{align}
\label{eq:19}
\tilde{L}_{nat}(t) = L_{nat}(t)/L_{nat}(0), ~\tilde{L}_{adv}(t) = L_{adv}(t)/L_{adv}(0), 
\end{align}
especially in our task, $L_{nat}(t)$ and $L_{adv}(t)$ represent the knowledge scales' gap for the student from the clean teacher $T_{nat}$ and robust teacher $T_{adv}$ at time $t$. So $\tilde{L}_{nat}(t) $ and $\tilde{L}_{adv}(t)$ can reflect the knowledge proportion that the student has not learned from the different teacher's knowledge at the time $t$ compared with time $0$. When $\tilde{L}_{nat}(t) $ and $\tilde{L}_{adv}(t)$ are equal, the student obtains relatively equal knowledge scales from different teachers, which is exactly what we search for. So our optimization goal is as follows:
\begin{align}
\label{eq:20}
\min||\tilde{L}_{nat}(t) - \tilde{L}_{adv}(t)||.
\end{align}

During the student's training process, the most direct impact on $\tilde{L}_{nat}(t) $ and $\tilde{L}_{adv}(t)$  is the loss weight $w_{nat}$ and $w_{adv}$. Intuitively speaking, when $\tilde{L}_{nat}(t)$ is smaller than $\tilde{L}_{adv}(t)$, the student has learned more knowledge scales from the clean teacher compared with the robust teacher during the 0 to t time, and the student is supposed to have smaller loss weight $w_{nat}$ in the learning process to limit the learning speed from this type of knowledge.

Based on the above criterion and analysis, we can dynamically balance  $\tilde{L}_{nat}(t)$ and $\tilde{L}_{adv}(t)$  by applying a relative weight $r_{nat}(t)$  and $r_{adv}(t)$ that we expected for loss at time t, which can be formulated as follows:
\begin{align}
\label{eq:21}
r_{nat}(t) = [\tilde{L}_{nat}(t)]^\beta/ \big[ [\tilde{L}_{nat}(t)]^\beta + [\tilde{L}_{adv}(t)]^\beta \big],
\end{align}
\begin{align}
\label{eq:22}
r_{adv}(t) = [\tilde{L}_{adv}(t)]^\beta/ \big[ [\tilde{L}_{nat}(t)]^\beta + [\tilde{L}_{adv}(t)]^\beta \big],
\end{align}
where $[\tilde{L}(t)]^\beta$ denotes the $\tilde{L}(t)$ power of $\beta$, and $\beta$ is set to control the strength of correcting the imbalance. The $r_{nat}(t)$ and $r_{adv}(t)$ strengthen the disadvantaged losses and weaken the advantaged losses to eliminate the imbalance between the student's learning speeds in different types of knowledge. The big $\beta$ is applicable when the vibration of relative loss value is obvious, while the small $\beta$ is suitable for $\tilde{L}_{nat}(t)$ and $\tilde{L}_{adv}(t)$ with stable influence abilities. 

From the consideration of training stability, we do not directly use $r_{nat}(t)$ and  $r_{adv}(t)$ as the weight of the loss, but update the original weight $w_{nat}(t)$ and $w_{adv}(t)$, which can be formulated as follows: 
\begin{align}
\label{eq:23}
w_{nat}(t) = r_{w}r_{nat}(t) + (1-r_{w}) w_{nat}(t-1),
\end{align}
\begin{align}
\label{eq:24}
w_{adv}(t) = r_{w}r_{adv}(t) + (1-r_{w}) w_{adv}(t-1).
\end{align}


On a practical level, with the assistance of the Normalization Loss Balance algorithm, the student can dynamically adjust the different loss weights in the whole training process according to the degree of acceptance of knowledge.  The student's learning speed will be inhibited and the corresponding loss weight will decrease if too much of this type of knowledge is accepted; On the contrary, the student's learning speed will be accelerated and the corresponding loss weight will increase if this type of knowledge is insufficient. Finally, the student tends to learn well from both two teachers, gaining relatively equal clean and robust knowledge, rather than appearing to have a partial ability. 

\subsection{Overview of B-MTARD}
All in all, to transfer both clean and robust knowledge to the student, we propose the B-MTARD to apply two teachers and propose the Entropy-Based Balance algorithm for teachers and the Normalization Loss Balance algorithm for the student to optimize the Adversarial Training process. The final minimization of B-MTARD is defined as follows: 
\begin{align}
\label{eq:25}
\begin{split}
    \mathop{\arg\min}_{\theta_S}&w_{nat}KL (S(x_{nat};\tau_{s}),T_{nat}(x_{nat};\tau_{nat}))  
    \\&+w_{adv} KL(S(x_{adv};\tau_{s}),T_{adv}(x_{adv};\tau_{adv})),
\end{split}
\end{align}
and the complete process of B-MTARD is in Algorithm \ref{algorithm:1}. 

\begin{algorithm}[t]  
  \caption{Overview of B-MTARD}  
  \label{algorithm:1}
  \begin{algorithmic}[1]   
   \Require {Initialize student $S$ with weight $\theta_S$ and temperature $\tau_{s}$, pretrained teacher $T_{nat}$ and $T_{adv}$, the initial temperature $\tau_{nat}$ and $\tau_{adv}$ for $T_{nat}$ and $T_{adv}$, respectively, the benign clean examples $x_{nat}$ and the labels $y$, the threat bound $\Omega$.}
         \For{$t = 0$ to $max$-$step$}  
            \State {\small Acquire adversarial example $ x_{adv}$ by Eq. (\ref{eq:5}).}
            \State {\small Get $L_{nat}(t)=KL(S(x_{nat};\tau_{s}),T_{nat}(x_{nat};\tau_{nat}))$.} 
            \State {\small Get $L_{adv}(t)=KL(S(x_{adv};\tau_{s}),T_{adv}(x_{adv};\tau_{adv}))$.} 
         	\If {$epoch = 0$ and $t = 0$}  
    			\State {\small Record $L_{nat}$ and $L_{adv}$ as $L_{nat}(0)$ and $L_{adv}(0)$.}  
    		\EndIf  
            \State{\small Update $w_{nat}(t)$ and $ w_{adv}(t)$ by Eq. (\ref{eq:23}) and Eq. (\ref{eq:24}).}
            \State{\small Update $\theta_S$ by Eq. (\ref{eq:25}).}
          \State{\small Update $\tau_{nat}$ and $\tau_{adv}$ by Eq. (\ref{eq:16}) and Eq. (\ref{eq:17}).}    
        \EndFor  
  \end{algorithmic}  
\end{algorithm}  

Compared with other existing methods, our method has several advantages. Firstly, our method can fit the different teacher-student combinations, which can be automatically adjusted by the Entropy-Based Balance algorithm and Normalization Loss Balance algorithm. Secondly, our method pays more attention to the overall performance based on Weighted Robust Accuracy, which measures the trade-off between accuracy and robustness. Thirdly, the training hyper-parameters can be dynamically updated by our adjustment algorithm and can fit the changes as training epochs increase, which is important to fit various scenarios.

\setlength{\tabcolsep}{1pt}
\begin{table*}[ht]
\begin{center}
\caption{Performance of the teacher networks used in our experiments.}
\label{table:1}
\scalebox{0.92}  { 
\begin{tabular}{m{2.5cm}<{\centering}m{2.5cm}<{\centering}m{1.3cm}<{\centering}m{1.3cm}<{\centering}m{1.3cm}<{\centering}m{1.5cm}<{\centering}m{1.3cm}<{\centering}m{1.3cm}<{\centering}m{1.3cm}<{\centering}}
\hline
Dataset & Model & Clean Acc & FGSM & PGD$_{\scriptscriptstyle \rm{sat}}$ & PGD$_{\scriptscriptstyle \rm{trades}}$ & CW$_{\scriptscriptstyle \rm{\infty}}$ & AA&Type\\
\hline
\multirow{2}*{CIFAR-10} & ResNet-56 & \textbf{93.18\%} &19.23\% &0\% &0\% &0\% &0\%& Clean\\
~ & WideResNet-34-10 & 84.91\% &\textbf{61.14\%} &\textbf{55.30\%} &\textbf{56.61\%} &\textbf{53.84\%}&\textbf{53.08\%}& Robust \\
\hline
\multirow{2}*{CIFAR-100} & WideResNet-22-6 & \textbf{76.65\%} &4.85\% &0\% &0\% &0\% &0\%& Clean \\
~ & WideResNet-70-16 & 60.96\% &\textbf{35.89\%} &\textbf{33.58\%} &\textbf{33.99\%} &\textbf{31.05\%} &\textbf{30.03\%} & Robust\\
\hline
\multirow{2}*{Tiny-ImageNet} & PreActResNet-34 & \textbf{66.05\%} &2.50\% &0.04\% &0.09\% &0\%&0\%& Clean \\
~ & PreActResNet-34 & 52.76\% &\textbf{27.05\%} &\textbf{24.00\%} &\textbf{24.63\%} &\textbf{20.07\%} &\textbf{18.92\%} & Robust\\
\hline
\end{tabular}
}
\end{center}
\end{table*}
\setlength{\tabcolsep}{1pt}
\begin{table*}[ht]
\caption{White-box robustness on CIFAR-10, CIFAR-100, and Tiny-ImageNet datasets. The results of CIFAR-10 and CIFAR-100 are based on the ResNet-18, and the results of Tiny-ImageNet are based on the PreActResNet-18. All results are the best checkpoint based on W-Robust Acc.}
\begin{center}
\label{table:2}
\scalebox{0.92}  { 
\begin{tabular}{m{2cm}<{\centering}m{2.4cm}<{\centering}|m{1.2cm}<{\centering}|m{1.2cm}<{\centering}m{1.4cm}<{\centering}|m{1.2cm}<{\centering}m{1.4cm}<{\centering}|m{1.2cm}<{\centering}m{1.4cm}<{\centering}|m{1.2cm}<{\centering}m{1.4cm}<{\centering}|m{1.2cm}<{\centering}m{1.4cm}<{\centering}}
\hline
~&~& ~&\multicolumn{2}{c|}{FGSM\cite{goodfellow2014explaining}} &  \multicolumn{2}{c|}{PGD$_{\scriptscriptstyle \rm{sat}}$\cite{madry2017towards}} &  \multicolumn{2}{c|}{PGD$_{\scriptscriptstyle \rm{trades}}$\cite{zhang2019theoretically}}  &  \multicolumn{2}{c|}{CW$_{\scriptscriptstyle \rm{\infty}}$\cite{carlini2017towards}} &  \multicolumn{2}{c}{AutoAttack\cite{croce2020reliable}}\\
\hline
 DataSet  &  Defense  & Clean & Robust& W-Robust & Robust& W-Robust& Robust& W-Robust&Robust& W-Robust&Robust& W-Robust\\
\hline
 \multirow{10}*{CIFAR-10} & Natural &94.57\% &18.60\% &56.59\% & 0\% & 47.29\% & 0\% & 47.29\%& 0\% & 47.29\%& 0\% & 47.29\% \\
 ~ & SAT\cite{madry2017towards} &84.20\% &55.59\% &69.90\% &45.95\% &65.08\% & 48.12\% &66.16\% &45.97\% &65.09\% & 43.49\% & 63.85\%\\
 ~ & TRADES\cite{zhang2019theoretically} &83.00\% &58.35\% &70.68\% &  52.35\% &67.68\% & 53.83\% & 68.42\% & 50.23\% & 66.62\% & 49.03\% & 66.02\%\\
 ~ & ARD\cite{goldblum2020adversarially} &84.11\% &58.40\% &71.26\% &  50.93\% & 67.52\% &52.96\%& 68.54\%&50.15\% & 67.13\% & 48.20\% & 66.16\%\\
 ~ & RSLAD\cite{zi2021revisiting} &83.99\% &60.41\% &72.20\% & 53.94\% & 68.97\% & 55.73\% & 69.86\% & 52.67\% &68.33\% & 50.98\% &67.49\%\\
 ~ & SCORE\cite{pang2022robustness} &84.43\% &59.84\% &72.14\% &  53.72\% & 69.08\% & 55.21\% & 69.82\% &  50.46\% & 67.45\% & 49.25\% & 66.84\%\\
 
 ~ & Fair-ARD\cite{Yue2023Revisiting} & 83.41\% & 58.91\% & 71.16\% & 52.00\% & 67.71\% & 53.77\% & 68.59\%&51.07\% & 67.24\% & 49.21\% & 66.31\% \\
 ~ & ABSLD\cite{zhao2023improving} & 83.21\% & 60.22\% & 71.72\% & 54.63\% & 68.92\% & 56.10\% & 69.66\% & 52.04\% & 67.63\% & 50.60\% & 66.91\% \\
  ~ & \textbf{MTARD}\cite{zhao2022enhanced} & 87.36\% & 61.20\% & 74.28\% & 50.73\% & 69.05\% & 53.60\% & 70.48\% & 48.57\% & 67.97\% & 46.18\% &  66.77\%  \\
 ~ & \textbf{B-MTARD (ours)} &88.20\% &61.42\% &\textbf{74.81\%} & 51.68\% & \textbf{69.94\%} & 54.40\%& \textbf{71.30\%}& 49.88\% &\textbf{69.04\%} &47.44\% &\textbf{67.82\%}  \\
\hline
 \multirow{10}*{CIFAR-100} & Natural & 75.18\%& 7.96\%& 41.57\%&0\%& 37.59\%&0\%& 37.59\%&0\%& 37.59\%&0\%& 37.59\%\\
~ & SAT\cite{madry2017towards} &56.16\% &25.88\%& 41.02\% & 21.18\%& 38.67\%& 22.02\%& 39.09\%&  20.90\%& 38.53\% &  19.76\% &  37.96\%\\
 ~ & TRADES\cite{zhang2019theoretically} &57.75\%& 31.36\% &44.56\% &  28.05\% &  42.90\% &  28.88\% &43.32\%&24.19\% &40.97\% & 23.26\%& 40.51\% \\
 ~ & ARD\cite{goldblum2020adversarially} &60.11\%& 33.61\% &46.86\% &  29.40\%& 44.76\%&  30.51\%& 45.31\%& 27.56\% &43.84\%&25.71\%& 42.91\% \\
 ~ & RSLAD\cite{zi2021revisiting} &58.25\%& 34.73\%& 46.49\%& 31.19\%& 44.72\%& 32.05\%& 45.15\% & 28.21\%& 43.23\% &  26.50\% &42.38\%\\
  ~ & SCORE\cite{pang2022robustness} &56.40\%& 32.94\%& 44.67\% & 30.27\% &43.34\%&30.56\% &43.48\% &  26.30\% &41.35\%& 24.74\% &40.57\%\\
 ~ & Fair-ARD\cite{Yue2023Revisiting} & 57.81\% & 34.39\% &46.10\% & 30.64\% & 44.23\% & 31.50\% & 44.66\% & 27.84\% & 42.83\% & 26.22\% & 42.02\% \\
 ~ & ABSLD\cite{zhao2023improving} & 56.77\% & 34.94\% & 45.86\% & 32.41\% & 44.59\% & 32.99\% & 44.88\% & 26.99\% & 41.88\% & 25.38\% & 41.08\% \\
  ~ & \textbf{MTARD}\cite{zhao2022enhanced} & 64.30\% &31.49\%& 47.90\% & 24.95\% & 44.63\%&26.75\%& 45.53\%  & 23.42\% &43.86\% & 21.31\% & 42.81\%\\
 ~ & \textbf{B-MTARD (ours)} & 65.08\%& 34.21\%&\textbf{49.65\%}& 28.50\% &\textbf{46.79\%} & 29.94\% &\textbf{47.51\%} &  25.45\% &\textbf{45.27\%}&  23.98\% & \textbf{44.53\%}\\
\hline
 \multirow{10}*{Tiny-ImageNet}&  Natural &65.03\%& 1.73\%& 33.38\% &  0.03\%& 32.53\% & 0.05\%& 32.54\% &  0\%& 32.52\% & 0\%& 32.52\%\\
 ~& SAT\cite{madry2017towards}&50.08\%& 25.35\% &37.72\%& 22.24\%& 36.16\%& 23.05\% &36.57\%& 20.48\%& 35.28\%&18.54\%& 34.31\%\\
 ~ & TRADES\cite{zhang2019theoretically} &48.45\%& 23.59\%& 36.02\% & 21.59\%& 35.02\%& 22.09\%& 35.27\% &  17.33\%& 32.89\% & 16.66\%& 32.56\%
\\
 ~ & ARD\cite{goldblum2020adversarially} &53.22\%& 27.97\%& 40.60\%& 24.92\%& 39.07\% & 25.71\%& 39.47\%& 21.41\%& 37.32\%& 19.91\%& 36.57\%\\
 ~ & RSLAD\cite{zi2021revisiting}  & 48.78\%& 27.26\%& 38.02\%& 25.00\%& 36.89\% & 25.45\%& 37.12\% & 20.87\%& 34.83\%& 17.11\%& 32.95\%
\\
  ~ & SCORE\cite{pang2022robustness} &10.05\% &7.80\%& 8.93\% & 7.65\%& 8.85\% & 7.67\% & 8.86\% &  6.19\% & 8.13\% & 5.97\%& 8.01\%
\\
 ~ & Fair-ARD\cite{Yue2023Revisiting} & 46.64\% & 25.81\% & 36.23\% & 23.91\% & 35.28\% & 24.29\% & 35.47\% & 19.59\% & 33.12\% & 17.51\% & 32.08\% \\
 ~ & ABSLD\cite{zhao2023improving} & 47.21\% & 25.79\% & 36.50\% & 23.59\% & 35.40\% & 23.98\% & 35.60\% & 19.50\% & 33.36\% & 17.84\% & 32.53\% \\
  ~ & \textbf{MTARD}\cite{zhao2022enhanced} & 52.98\% & 26.41\% & 39.70\% &22.55\% & 37.77\% & 23.41\% & 38.20\% & 19.36\% & 36.17\% & 16.84\% & 34.91\%   \\  
 ~ & \textbf{B-MTARD (ours)} &56.81\%& 28.12\%& \textbf{42.47\%} &  23.93\%& \textbf{40.37\%}& 24.94\%& \textbf{40.88\%}& 19.69\% &\textbf{38.25\%} & 17.71\%& \textbf{37.26\%}\\
\hline
\end{tabular}
}
\end{center}
\end{table*}
\begin{table*}[t]
\begin{center}
\caption{White-box robustness on CIFAR-10, CIFAR-100, and Tiny-ImageNet datasets. The results of three datasets are trained based on the MobileNet-v2. All results are the best checkpoint based on W-Robust Acc.}
\label{table:3}
\scalebox{0.92}  { 
\begin{tabular}{m{2cm}<{\centering}m{2.4cm}<{\centering}|m{1.2cm}<{\centering}|m{1.2cm}<{\centering}m{1.4cm}<{\centering}|m{1.2cm}<{\centering}m{1.4cm}<{\centering}|m{1.2cm}<{\centering}m{1.4cm}<{\centering}|m{1.2cm}<{\centering}m{1.4cm}<{\centering}|m{1.2cm}<{\centering}m{1.4cm}<{\centering}}
\hline
~&~& ~&\multicolumn{2}{c|}{FGSM\cite{goodfellow2014explaining}} &  \multicolumn{2}{c|}{PGD$_{\scriptscriptstyle \rm{sat}}$\cite{madry2017towards}} &  \multicolumn{2}{c|}{PGD$_{\scriptscriptstyle \rm{trades}}$\cite{zhang2019theoretically}}  &  \multicolumn{2}{c|}{CW$_{\scriptscriptstyle \rm{\infty}}$\cite{carlini2017towards}} &  \multicolumn{2}{c}{AutoAttack\cite{croce2020reliable}}\\
\hline
 DataSet  &  Defense  & Clean & Robust& W-Robust & Robust& W-Robust& Robust& W-Robust&Robust& W-Robust&Robust& W-Robust\\
\hline
 \multirow{10}*{CIFAR-10} & Natural&93.35\%& 12.22\%& 52.79\% & 0\%& 46.68\% & 0\%& 46.68\% & 0\%& 46.68\% & 0\%& 46.68\%   \\
 ~ & SAT\cite{madry2017towards} & 83.87\%& 55.89\%& 69.88\% & 46.84\%& 65.36\%& 49.14\%& 66.51\% &  46.62\%& 65.25\% & 43.43\%& 63.65\%\\
 ~ & TRADES\cite{zhang2019theoretically}&77.95\%& 53.75\%& 65.85\% & 49.06\%& 63.51\%& 50.27\%& 64.11\%& 46.06\%& 62.01\% & 45.18\%& 61.66\%\\
 ~ & ARD\cite{goldblum2020adversarially} &83.43\%& 57.03\%& 70.23\%&49.50\%& 66.47\%& 51.70\%& 67.57\%& 48.96\%& 66.20\% &  46.60\%& 65.02\%\\
 ~ & RSLAD\cite{zi2021revisiting}& 83.20\%& 59.47\%& 71.34\%&53.25\%& 68.23\%&54.76\%& 68.98\% & 51.78\% & 67.49\% & 50.23\%& 66.72\%\\
 ~ & SCORE\cite{pang2022robustness} & 82.32\%& 58.43\%& 70.38\%&53.42\%& 67.87\%& 54.46\%& 68.39\% & 49.18\% & 65.75\% & 48.39\%& 65.36\%\\
  
 ~ & Fair-ARD\cite{Yue2023Revisiting} & 82.65\% & 56.37\% & 69.51\% & 50.50\% & 66.58\% & 52.12\% & 67.39\% &  51.07\% & 66.86\% & 47.68\% & 65.17\% \\
 ~ & ABSLD\cite{zhao2023improving} & 82.50\% & 58.47\% & 70.49\% & 52.98\% & 67.74\% & 54.49\% & 68.50\% & 50.20\% & 66.35\% & 48.65\% & 65.58\% \\
  ~ & \textbf{MTARD}\cite{zhao2022enhanced} & 89.26\%& 57.84\%& 73.55\% &44.16\%& 66.71\%& 47.99\%& 68.63\% & 43.42\%& 66.34\% & 41.02\% & 65.14\%\\
 ~ & \textbf{B-MTARD (ours)} & 89.09\% & 58.79\% & \textbf{73.94\%} & 47.56\% & \textbf{68.33\%} & 50.44\% & \textbf{69.77\%} &  46.81\% & \textbf{67.95\%} & 44.58\% & \textbf{66.84\%}  \\
\hline
 \multirow{10}*{CIFAR-100} & Natural &  74.86\%& 5.94\% & 40.40\% &0\%& 37.43\% &0\%& 37.43\% &0\%& 37.43\%&0\%& 37.43\% \\
~ & SAT\cite{madry2017towards} & 59.19\% & 30.88\%& 45.04\% & 25.64\% &42.42\%& 26.96\%& 43.08\%& 25.01\%& 42.10\%& 22.84\%& 41.02\%\\
 ~ & TRADES\cite{zhang2019theoretically} & 55.41\%& 30.28\% &42.85\% & 23.33\%& 39.37\% &  28.42\%& 41.92\% &  27.72\%& 41.57\%& 22.61\%& 39.01\%\\
 ~ & ARD\cite{goldblum2020adversarially} &60.45\%& 32.77\%& 46.61\% & 28.69\%& 44.57\%&29.63\%& 45.04\% & 26.55\%& 43.50\%& 24.86\%& 42.66\% \\
 ~ & RSLAD\cite{zi2021revisiting} &59.01\%& 33.88\%& 46.45\%& 30.19\%& 44.60\%&31.19\%& 45.10\%& 27.98\%& 43.50\%&26.33\%& 42.67\%\\
  ~ & SCORE\cite{pang2022robustness}&49.38\%& 29.28\%& 39.33\% & 27.03\%& 38.21\%& 27.53\%& 38.46\%&  23.29\%& 36.34\%& 21.92\%& 35.65\%\\
   
 ~ & Fair-ARD\cite{Yue2023Revisiting} & 59.18\% & 34.07\% & 46.63\% & 30.15\% & 44.67\% & 31.26\% & 45.22\% & 27.55\% & 43.37\% & 25.84\% & 42.51\% \\
 ~ & ABSLD\cite{zhao2023improving} & 56.67\% & 33.85\% & 45.26\% & 31.28\% & 43.98\% & 31.90\% & 44.29\% & 26.40\% & 41.54\% & 24.56\% & 40.62\% \\
  ~ & \textbf{MTARD}\cite{zhao2022enhanced} & 67.01\%& 32.42\%& 49.72\%&25.14\%& 46.08\%&27.10\%& 47.06\%&24.14\%& 45.58\%&21.61\%& 44.31\%\\
 ~ & \textbf{B-MTARD (ours)}&66.13\%& 34.36\%&\textbf{50.25\%} &28.47\%& \textbf{47.30\%}&29.82\%& \textbf{47.98\%}& 26.50\%&\textbf{46.32\%}& 24.01\%& \textbf{45.07\%} \\
\hline
 \multirow{10}*{Tiny-ImageNet}&  Natural &62.09\%& 0.72\%& 31.41\% & 0\%& 31.05\% & 0\%& 31.05\% & 0\%& 31.05\%& 0\%& 31.05\%\\
 ~& SAT\cite{madry2017towards}&49.03\%& 23.38\%& 36.21\%& 20.31\%& 34.67\%&21.15\%& 35.09\%&18.69\%& 33.86\%&17.35\%& 33.19\%\\
 ~ & TRADES\cite{zhang2019theoretically} & 43.81\%& 20.10\%& 31.96\% & 18.16\%& 30.99\%&18.36\%& 31.09\%&13.47\%& 28.66\%& 12.82\%& 28.32\%
\\
 ~ & ARD\cite{goldblum2020adversarially} &45.53\%& 22.88\%& 33.21\% & 20.43\%& 32.98\%&21.00\%& 33.27\%& 16.81\%& 31.17\%&15.40\%& 30.47\%
\\
 ~ & RSLAD\cite{zi2021revisiting}  & 45.69\%& 24.09\%& 34.89\%& 22.30\%& 34.00\%&22.74\%& 34.22\%& 18.63\%& 32.16\%&17.09\%& 31.39\%
\\
  ~ & SCORE\cite{pang2022robustness} &28.27\%& 17.47\%& 22.87\%& 16.48\%& 22.38\%& 16.69\%& 22.48\%&13.25\%& 20.76\%&12.30\%& 20.29\%
\\
 
 ~ & Fair-ARD\cite{Yue2023Revisiting} & 47.24\% & 25.31\% & 36.28\% & 23.37\% & 35.31\% & 23.77\% & 35.51\% & 20.04\% & 33.64\% & 18.09\% & 32.67\%  \\
 ~ & ABSLD\cite{zhao2023improving} & 48.08\% & 26.21\% & 37.15\% & 26.21\% & 37.15\% & 24.39\% & 36.24\% & 20.10\% & 34.09\% & 18.38\% & 33.23\% \\
~ & \textbf{MTARD}\cite{zhao2022enhanced} & 50.50\% & 23.94\% & 37.22\% & 20.45\% & 35.48\% & 21.20\% & 35.85\% & 17.45\% &  33.98\% & 15.06\% & 32.78\%   \\
 ~ & \textbf{B-MTARD (ours)} &52.98\%& 25.60\%& \textbf{39.29\%}&21.58\%& \textbf{37.28\%}&22.58\%& \textbf{37.78\%}& 18.08\%& \textbf{35.53\%}& 15.68\%& \textbf{34.33\%}\\
\hline
\end{tabular}
}
\end{center}
\end{table*}

\section{Experiments}
\label{sec:Experiment}
In this section, we initially describe the experimental setting and then evaluate the Weighted Robust Accuracy of our B-MTARD and several baseline methods under prevailing white-box attacks and black-box attacks including transfer-based and query-based attacks. We also conduct a series of ablation studies to demonstrate the effectiveness. 
\subsection{Experimental Settings}
We conduct experiments on three datasets including CIFAR-10 \cite{krizhevsky2009learning}, CIFAR-100, and Tiny-ImageNet \cite{le2015tiny}. We apply the standard training method and several state-of-the-art methods for comparison: Adversarial Training method: SAT \cite{madry2017towards}, Adversarial Robustness Distillation methods: ARD \cite{goldblum2020adversarially}, RSLAD \cite{zi2021revisiting}, {Fair-ARD \cite{Yue2023Revisiting}, and ABSLD \cite{zhao2023improving}}; Mitigating Trade-off methods: TRADES \cite{zhang2019theoretically} and SCORE \cite{pang2022robustness}. 
{Meanwhile, we also show the results of our conference version for comparison: MTARD \cite{zhao2022enhanced}.}

\textbf{Student and Teacher Networks. } Here we consider two student networks for CIFAR-10 and CIFAR-100 including ResNet-18 \cite{he2016deep} and MobileNet-v2 \cite{sandler2018mobilenetv2}. For Tiny-ImageNet, we select PreActResNet-18 \cite{he2016identity} and MobileNet-v2 following previous work. As for teacher models, we choose clean teacher networks including ResNet-56 for CIFAR-10,  WideResNet-22-6 \cite{zagoruyko2016wide} for CIFAR-100, and PreActResNet-34 \cite{he2016identity} for Tiny-ImageNet. Three robust teacher networks include WideResNet-34-10 for CIFAR-10, WideResNet-70-16 \cite{gowal2020uncovering} for CIFAR-100, and PreActResNet-34 for Tiny-ImageNet. For CIFAR-10, WideResNet-34-10 is trained using TRADES \cite{zhang2019theoretically}; For CIFAR-100, we use the WideResNet-70-16 model provided by Gowal et al. \cite{gowal2020uncovering}; In Addition, for Tiny-ImageNet, we use  PreActResNet-34 trained by TRADES \cite{zhang2019theoretically} as the robust teacher. The teachers are pre-trained and will not be changed during the training process. The performance of these teacher models is shown in Table \ref{table:1}.

\textbf{Training Setting.} We train the student using Stochastic Gradient Descent (SGD) optimizer with an initial learning rate of 0.1, a momentum of 0.9, and a weight decay of 2e-4. For B-MTARD, the weight loss learning rate $r_w$ is initially set as 0.025; the temperature learning rate $r_{\tau}$ is initially set as 0.001. the student's temperature $\tau_{s}$ is set to constant 1. Meanwhile, the $\tau_{nat}$ and $\tau_{adv}$ are initially set as 1 without additonal instruction, {and the $\tau_{nat}$ and $\tau_{adv}$ are initially set as 2 for MobileNet-v2 on CIFAR-10. 
The maximum and minimum values of temperature are 10 and 1,  respectively.} For CIFAR-10 and CIFAR-100, we set the total number of training epochs to 300. The learning rate is divided by 10 at the 215-th, 260-th, and 285-th epochs; For Tiny-ImageNet, we set the total number of training epochs to 100, and the learning rate is divided by 10 at the 75-th and 90-th epochs. We set the batch size to 128 (CIFAR-10 and CIFAR-100) and 32 (Tiny-ImageNet), and $\beta$ is set to 1. For the inner maximization of B-MTARD, we use a 10-step PGD with a random start size of 0.001 and a step size of 2/255. All training perturbation in the maximization is bounded to the $L_{\infty}$ norm $\epsilon$ = 8/255. For the training of the natural model, we train the networks for 100 epochs, and the learning rate is divided by 10 at the 75-th and 90-th epochs. 

For the comparison methods, we follow SAT, TRADES, SCORE, ARD, RSLAD, Fair-ARD, and ABSLD original settings without additional instruction. The SCORE version is the TRADES+LSE without additional data. For ARD, we use the same robust teacher as RSLAD and MTARD. The temperature $\tau$ of ARD is set to 30 on CIFAR-10 while set to 5 on CIFAR-100 and Tiny-ImageNet, and the $\alpha$ is set to 0.95 following \cite{goldblum2020adversarially} on CIFAR-100 and Tiny-ImageNet. The learning rate of SAT and SCORE for Tiny-ImageNet is 0.01 to achieve better performance.

\textbf{Evaluation Setting. } The same as previous studies, we evaluate the models against white box adversarial attacks: FGSM \cite{goodfellow2014explaining}, PGD$_{\scriptscriptstyle \rm{sat}}$\cite{madry2017towards}, PGD$_{\scriptscriptstyle \rm{trades}}$\cite{zhang2019theoretically}, CW$_{\scriptscriptstyle \rm{\infty}}$\cite{carlini2017towards}, which are commonly used adversarial attacks in adversarial robustness evaluation. Meanwhile, we also apply a strong attack: AutoAttack (AA) \cite{croce2020reliable} {(an evaluation tool in RobustBench \cite{croce2021robustbench})} to evaluate the robustness.  The step sizes of PGD$_{\scriptscriptstyle \rm{sat}}$ and PGD$_{\scriptscriptstyle \rm{trades}}$ are 2/255 and 0.003, respectively, and the step is 20. The total step of CW$_{\scriptscriptstyle \rm{\infty}}$ is 30. Maximum perturbation is bounded to the $L_{\infty}$ norm $\epsilon$ = 8/255 for all attacks. 
Meanwhile, we conduct a black-box evaluation, which includes the transfer-based attack and query-based attack to test the robustness in a near-real environment. As for the transfer-based attack, we choose  robust teachers (WideResNet-34-10 for CIFAR-10, WideResNet-70-16 for CIFAR-100, and PreActResNet-34 for Tiny-ImageNet) as the surrogate models to produce adversarial example against the PGD$_{\scriptscriptstyle \rm{trades}}$ and CW$_{\scriptscriptstyle \rm{\infty}}$ attack; As for the query-based attack, we choose a score-based attack: Square Attack \cite{andriushchenko2020square}  {and a decision-based attack: RayS Attack \cite{chen2020rays}}, and the queries are 100.
\setlength{\tabcolsep}{1pt}
\begin{figure}[t]
  \centering
  \includegraphics[width=0.9\linewidth]{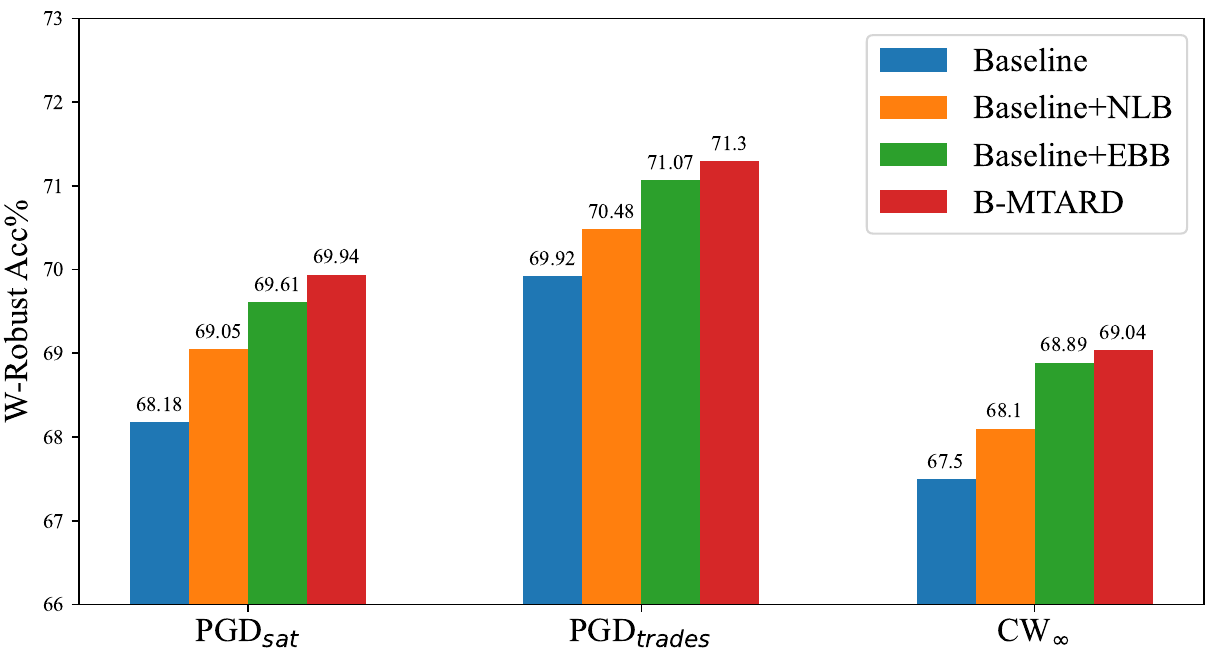}\\
\caption{Ablation study with ResNet-18 student trained using variants of our B-MTARD and Baseline method on CIFAR-10. NLB and EBB represent Normalization Loss Balance (NLB) and Entropy-Based Balance (EBB). B-MTARD is our final version, which represents Baseline+NLB+EBB. {The Baseline+NLB is the result in our ECCV version \cite{zhao2022enhanced}.} All the results are the best checkpoint based on W-Robust Acc.}
\label{fig:result1}
\end{figure}
\begin{figure}[t]
  \centering
    \includegraphics[width=0.48\linewidth]{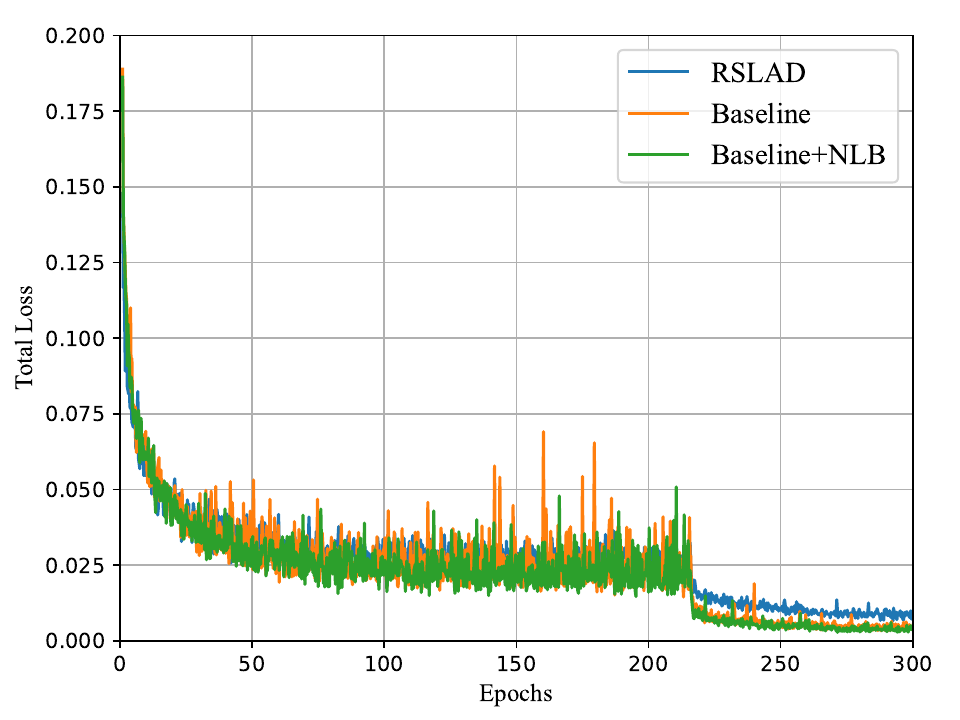}
  \includegraphics[width=0.48\linewidth]{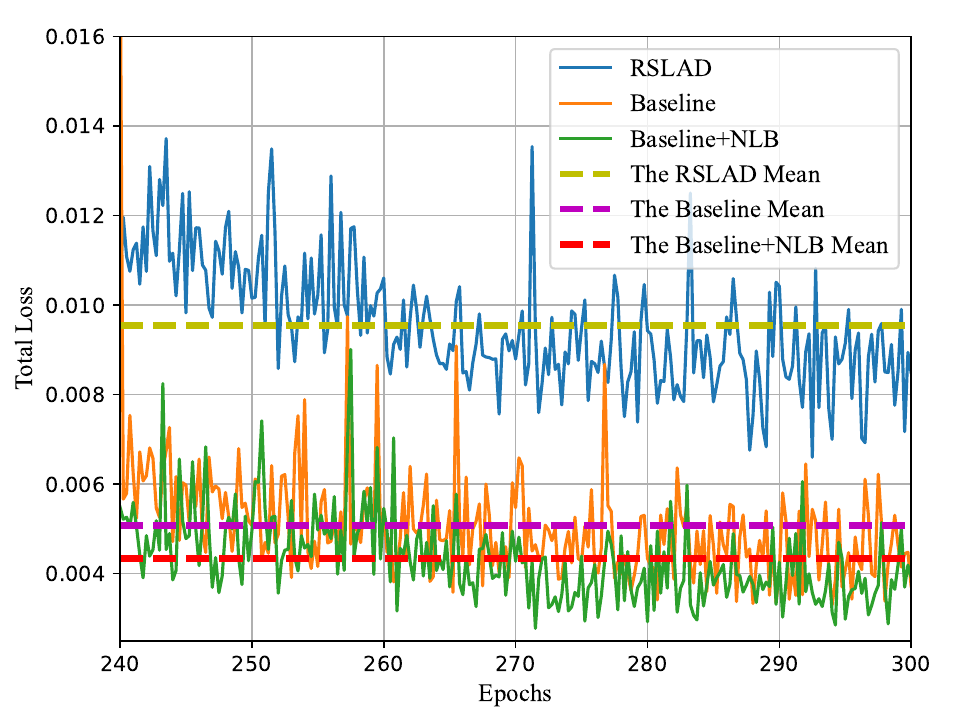}\\
\caption{The total loss $L_{total}$ with the ResNet-18 student trained using variants of RSLAD, Baseline, and Baseline+NLB on CIFAR-10. NLB is abbreviation of Normalization Loss Balance.  The y-axis is the $L_{total}$ in the training epoch x. The left is the change curve of $L_{total}$ in the whole training process, the right is the curve of $L_{total}$ in the final 60 epochs.}
\label{fig:result2}
\end{figure}

\begin{table*}[t]
\begin{center}
\caption{Black-box robustness on CIFAR-10, CIFAR-100, and Tiny-ImageNet datasets. The results of CIFAR-10 and CIFAR-100 are based on the ResNet-18, and the results of Tiny-ImageNet are based on the PreActResNet-18. All results are the best checkpoint based on W-Robust Acc.}
\label{table:4}
\scalebox{0.92}  { 
\begin{tabular}{m{2cm}<{\centering}m{2.4cm}<{\centering}|m{1.2cm}<{\centering}|m{1.2cm}<{\centering}m{1.4cm}<{\centering}|m{1.2cm}<{\centering}m{1.4cm}<{\centering}|m{1.2cm}<{\centering}m{1.4cm}<{\centering}|m{1.2cm}<{\centering}m{1.4cm}<{\centering}}
\hline
~&~& ~&\multicolumn{2}{c|}{PGD$_{\scriptscriptstyle \rm{trades}}$\cite{zhang2019theoretically}} &  \multicolumn{2}{c|}{CW$_{\scriptscriptstyle \rm{\infty}}$\cite{carlini2017towards}} &  \multicolumn{2}{c|}{Square Attack\cite{andriushchenko2020square}}  &  \multicolumn{2}{c}{RayS Attack\cite{chen2020rays}} \\
\hline
 DataSet  &  Defense  & Clean & Robust& W-Robust & Robust& W-Robust& Robust& W-Robust&Robust& W-Robust\\
\hline
 \multirow{9}*{CIFAR-10} &  SAT\cite{madry2017towards} & 84.20\%& 64.84\%& 74.52\%& 63.84\%& 74.02\% &  72.48\%& 78.34\% & 74.13\% & 79.17\% \\
 ~ & TRADES\cite{zhang2019theoretically}&83.00\%& 63.61\%& 73.31\%&62.83\%& 72.92\% &72.49\%& 77.75\% & 73.59\% & 78.30\% \\
 ~ & ARD\cite{goldblum2020adversarially} &84.11\%& 63.50\%& 73.81\%&62.86\%& 73.49\%&74.60\%& 79.36\% & 75.92\% & 80.02\%  \\
 ~ & RSLAD\cite{zi2021revisiting}& 83.99\%& 63.96\%& 73.98\%& 63.05\%& 73.52\% & 72.47\%& 78.23\% & 75.60\% & 79.80\% \\
 ~ & SCORE\cite{pang2022robustness} & 84.43\%& 64.34\%& 74.39\%&63.79\%& 74.11\% &  74.68\%& 79.56\% &75.87\% & 80.15\% \\
 ~ & Fair-ARD\cite{Yue2023Revisiting} & 83.41\% & 63.31\% & 73.36\% & 62.76\% & 73.09\% & 74.08\% & 78.75\% & 75.38\% & 79.40\% \\
 ~ & ABSLD\cite{zhao2023improving} & 83.21\% & 62.91\% & 73.06\% & 62.19\% & 72.70\% & 73.93\% &78.57\% & 74.90\% & 79.06\% \\
  ~ & \textbf{MTARD}\cite{zhao2022enhanced} & 87.36\% &65.26\% & 76.31\% & 64.58\% & 75.97\% & 78.58\% & 82.97\% & 80.13\% & 83.75\%  \\
 ~ & \textbf{B-MTARD (ours)} & 88.20\%& 65.29\%& \textbf{76.75\%} & 64.64\%& \textbf{76.42\%} & 79.82\%&\textbf{84.01\%} &80.98\% & \textbf{84.59\%}  \\
\hline
 \multirow{9}*{CIFAR-100} & SAT\cite{madry2017towards} & 56.16\%& 38.10\% &47.13\% & 39.42\%& 47.79\% & 40.05\%& 48.11\%& 40.92\% & 48.54\% \\
 ~ & TRADES\cite{zhang2019theoretically} & 57.75\% &38.20\%& 47.98\%& 38.63\%& 48.19\%&42.51\%& 50.13\% & 42.77\% & 50.26\% \\
 ~ & ARD\cite{goldblum2020adversarially} &60.11\%& 39.53\%& 49.82\% & 38.85\%& 49.48\%& 47.20\%& 53.66\% & 48.04\% & 56.08\% \\
 ~ & RSLAD\cite{zi2021revisiting} &58.25\%& 39.93\%& 49.09\% & 39.67\%& 48.96\%& 45.32\%& 51.79\% & 45.77\% & 52.01\% \\
  ~ & SCORE\cite{pang2022robustness}&56.40\%& 39.09\% &47.75\%& 40.02\%& 48.21\%&  43.98\% &50.34\%  & 44.57\% & 50.49\% \\
 ~ & Fair-ARD\cite{Yue2023Revisiting} & 57.81\% & 39.56\% & 48.69\% & 39.10\% & 48.46\% & 44.99\% & 51.40\% & 45.54\% & 51.68\% \\
 ~ & ABSLD\cite{zhao2023improving} & 56.77\% & 38.42\% & 47.60\% & 38.11\% & 47.44\% & 44.09\% & 50.43\% & 44.57\% & 50.67\% \\
  ~ & \textbf{MTARD}\cite{zhao2022enhanced} & 64.30\% & 41.46\% & 52.88\% & 41.18\% & 52.74\% & 48.13\% & 56.22\% & 49.73\% & 57.02\% \\
 ~ & \textbf{B-MTARD (ours)}&65.08\% &42.11\%& \textbf{53.60\%} &41.35\%& \textbf{53.22\%} &  49.40\%& \textbf{57.24\%} & 50.66\% & \textbf{57.87\%}  \\
\hline
 \multirow{9}*{Tiny-ImageNet}& SAT\cite{madry2017towards}&50.08\%& 33.40\%& 41.74\% & 33.20\%& 41.63\% & 38.72\% & 44.40\% & 39.35\% & 44.72\% \\
 ~ & TRADES\cite{zhang2019theoretically} & 48.45\%& 31.01\%& 39.73\% & 30.72\%& 39.59\%& 36.58\%& 42.52\% & 37.22\% & 42.84\%
\\
 ~ & ARD\cite{goldblum2020adversarially} &53.22\%& 34.74\%& 43.98\% & 33.32\%& 43.27\% & 42.58\% &47.90\% & 42.87\% & 48.05\%
\\
 ~ & RSLAD\cite{zi2021revisiting}  & 48.78\%& 32.85\%& 40.82\% &  32.09\% &  40.44\%& 37.64\%& 43.21\% & 38.15\% & 43.47\%
\\
  ~ & SCORE\cite{pang2022robustness} &10.05\%& 8.74\% &9.40\% &  8.82\% &  9.44\% &  8.67\% & 9.36\% & 8.70\% & 9.38\%
\\
 ~ & Fair-ARD\cite{Yue2023Revisiting} & 46.64\% & 31.58\% & 39.11\% & 31.38\% & 39.01\% & 35.81\% & 41.23\% & 36.44\% & 41.54\% \\
 ~ & ABSLD\cite{zhao2023improving} & 47.21\% & 31.84\% & 39.53\% & 31.66\% & 39.44\% & 36.77\% & 41.99\% & 37.42\% & 42.32\%  \\
  ~ & \textbf{MTARD}\cite{zhao2022enhanced} &  52.98\% & 34.48\% & 43.73\% & 33.80\% & 43.39\% & 41.70\% & 47.34\% & 42.55\% & 47.77\%  \\
 ~ & \textbf{B-MTARD (ours)} &56.81\%& 36.65\%& \textbf{46.73\%} & 33.40\% &  \textbf{45.11\%} &  44.46\% &\textbf{50.64\%} & 45.51\% & \textbf{51.16\%}
 \\
\hline
\end{tabular}
}
\end{center}
\end{table*}
\begin{table*}[ht] 
\begin{center}
\caption{Black-box robustness on CIFAR-10, CIFAR-100, and Tiny-ImageNet datasets. The results of three datasets are trained based on the MobileNet-v2. All results are the best checkpoint based on W-Robust Acc.}
\label{table:5}
\scalebox{0.92}  { 
\begin{tabular}{m{2cm}<{\centering}m{2.4cm}<{\centering}|m{1.2cm}<{\centering}|m{1.2cm}<{\centering}m{1.4cm}<{\centering}|m{1.2cm}<{\centering}m{1.4cm}<{\centering}|m{1.2cm}<{\centering}m{1.4cm}<{\centering}|m{1.2cm}<{\centering}m{1.4cm}<{\centering}}
\hline
~&~& ~&\multicolumn{2}{c|}{PGD$_{\scriptscriptstyle \rm{trades}}$\cite{zhang2019theoretically}} &  \multicolumn{2}{c|}{CW$_{\scriptscriptstyle \rm{\infty}}$\cite{carlini2017towards}} &  \multicolumn{2}{c|}{Square Attack\cite{andriushchenko2020square}}  &  \multicolumn{2}{c}{RayS Attack\cite{chen2020rays}} \\
\hline
 DataSet  &  Defense  & Clean & Robust& W-Robust & Robust& W-Robust& Robust& W-Robust&Robust& W-Robust\\
\hline
 \multirow{9}*{CIFAR-10} & SAT\cite{madry2017towards} &83.87\%& 64.66\%& 74.27\% &  64.24\%& 74.06\%& 73.01\% & 78.44\% & 74.72\% & 79.30\% \\
 ~ & TRADES\cite{zhang2019theoretically}&77.95\% &61.04\%& 69.50\%& 60.66\%& 69.31\% & 67.43\%& 72.69\% & 68.91\% & 73.43\%  \\
 ~ & ARD\cite{goldblum2020adversarially} &83.43\%& 63.28\%& 73.36\%&  62.83\%& 73.13\%&73.26\%& 78.35\% & 74.62\% & 79.03\% \\
 ~ & RSLAD\cite{zi2021revisiting}& 83.20\%& 64.33\%& 73.77\%&63.45\%& 73.33\% &73.07\%& 78.14\% & 74.49\% &  78.85\% \\
 ~ & SCORE\cite{pang2022robustness} & 82.32\%& 63.84\%& 73.08\% &63.03\%& 72.68\%& 72.41\%& 77.37\% & 73.45\% & 77.89\%   \\
 ~ & Fair-ARD\cite{Yue2023Revisiting} & 82.65\% &  62.62\% & 72.64\% & 62.00\% & 72.32\% & 72.44\% & 77.55\% & 73.50\% & 78.08\% \\
 ~ & ABSLD\cite{zhao2023improving} & 82.50\% & 62.91\% & 72.71\% & 61.87\% & 72.19\% & 72.05\% & 77.28\% & 73.29\% & 77.90\%  \\
  ~ & \textbf{MTARD}\cite{zhao2022enhanced} & 89.26\% & 66.30\% & 77.78\% & 65.68\% & 77.47\% & 79.13\% &84.20\% & 80.91\% & 85.09\%  \\
 ~ & \textbf{B-MTARD (ours)} & 89.09\% & 66.47\% & \textbf{77.78\%} & 65.96\% & \textbf{77.53\%} & 79.61\% & \textbf{84.35\%} &  81.14\% & \textbf{85.12\%} \\
\hline
 \multirow{9}*{CIFAR-100} & SAT\cite{madry2017towards} &59.19\%& 40.70\%& 49.95\%&40.97\% &50.08\% & 44.51\%& 51.85\% & 45.59\% & 52.39\% \\
 ~ & TRADES\cite{zhang2019theoretically} &55.41\%& 37.76\%& 46.59\%&38.02\%& 46.72\% & 40.71\%& 48.06\% & 41.23\% & 48.32\%  \\
 ~ & ARD\cite{goldblum2020adversarially} &60.45\%& 39.15\%& 49.80\%&38.53\%& 49.49\% & 46.95\%& 53.70\% & 47.91\% & 54.18\%   \\
 ~ & RSLAD\cite{zi2021revisiting} &59.01\%& 40.32\%& 49.67\% &  39.92\% &49.47\%& 45.66\%& 52.34\% & 46.33\% & 52.67\% \\
  ~ & SCORE\cite{pang2022robustness}&49.38\%& 40.00\%& 44.69\% &  36.91\% &43.15\%&  37.94\%& 43.66\% & 38.04\% & 43.71\% \\
 ~ & Fair-ARD\cite{Yue2023Revisiting} &59.18\% &40.55\% & 49.87\% & 40.14\% & 49.66\% & 45.29\% & 52.24\% & 46.00\% & 52.59\% \\
 ~ & ABSLD\cite{zhao2023improving} & 56.67\% & 38.08\% & 47.38\% & 38.18\% & 47.43\% & 43.23\% & 49.95\%  & 43.60\% & 50.14\% \\
  ~ & \textbf{MTARD}\cite{zhao2022enhanced} & 67.01\%& 43.23\% &\textbf{55.12\%} & 42.92\% & \textbf{54.97\%} & 50.64\% & \textbf{58.83\%} & 52.49\% & \textbf{59.75\%} \\
 ~ & \textbf{B-MTARD (ours)}&66.13\% &42.67\%& 54.40\% & 42.04\%& 54.09\% &  50.83\%& 58.48\% & 51.84\% &  58.99\%  \\
\hline
 \multirow{9}*{Tiny-ImageNet}& SAT\cite{madry2017towards}&49.03\%& 33.47\%& 41.25\% &  33.13\%& 41.08\%& 37.95\%& 43.49\%& 38.32\% & 43.68\%   \\
 ~ & TRADES\cite{zhang2019theoretically} & 43.81\%& 28.35\%& 36.08\%& 28.64\%& 36.23\%& 32.39\%& 38.10\% & 32.37\% & 38.09\%
\\
 ~ & ARD\cite{goldblum2020adversarially} &45.53\%& 30.73\%& 38.13\% & 30.23\%& 37.88\%&34.60\%& 40.07\% & 35.13\% & 40.33\%
\\
 ~ & RSLAD\cite{zi2021revisiting}  & 45.69\%& 31.20\%& 38.45\% & 31.10\%& 38.40\%&35.18\%& 40.44\% &  35.57\% & 40.63\%
\\
  ~ & SCORE\cite{pang2022robustness} &28.27\%& 21.82\%& 25.05\% & 22.19\%& 25.23\%& 22.16\%& 25.22\% & 22.41\% & 25.34\%
\\
 ~ & Fair-ARD\cite{Yue2023Revisiting} & 47.24\% & 31.80\% & 39.52\% & 31.40\% & 39.32\% & 36.53\% & 41.89\% & 37.22\% & 42.23\% \\
 ~ & ABSLD\cite{zhao2023improving}& 48.08\% & 32.89\% & 40.49\% & 32.43\% & 40.26\% & 38.18\% & 43.13\% &  38.48\% & 43.28\% \\
  ~ & \textbf{MTARD}\cite{zhao2022enhanced} & 50.50\% & 32.75\% & 41.63\% & 32.05\% & 41.28\% & 38.88\% &  44.69\%  & 40.01\% & 45.26\% \\
 ~ & \textbf{B-MTARD (ours)} &52.98\%& 34.25\%& \textbf{43.62\%}& 33.50\%& \textbf{43.24\%}& 40.62\%& \textbf{46.80\%} & 41.44\% & \textbf{47.21\%}
\\
\hline
\end{tabular}
}
\end{center}
\end{table*}
Here, we use Weighted Robust Accuracy (W-Robust Acc) \cite{gurel2021knowledge} to evaluate the trade-off between the clean and robust accuracy of the model, it is defined as follows:
\begin{align}
\label{eq:26}
\mathcal{A}_{f} = \pi_{{nat}}P_{{nat}}[f(x)=y] +\pi_{{adv}}P_{{adv}}[f(x)=y],
\end{align}
where W-Robust Acc $\mathcal{A}_{f}$ are the accuracy of a model $f$ on x drawn from either the clean distribution $P_{nat}$ and the adversarial distribution $P_{adv}$. We set $\pi_{{nat}}$ and $\pi_{{adv}}$ both to 0.5, which means accuracy and robustness are equally important for comprehensive performance in the model. 

\setlength{\tabcolsep}{1.4pt}
\subsection{Ablation Study}
\begin{figure}[t]
  \centering
    \includegraphics[width=0.48\linewidth]{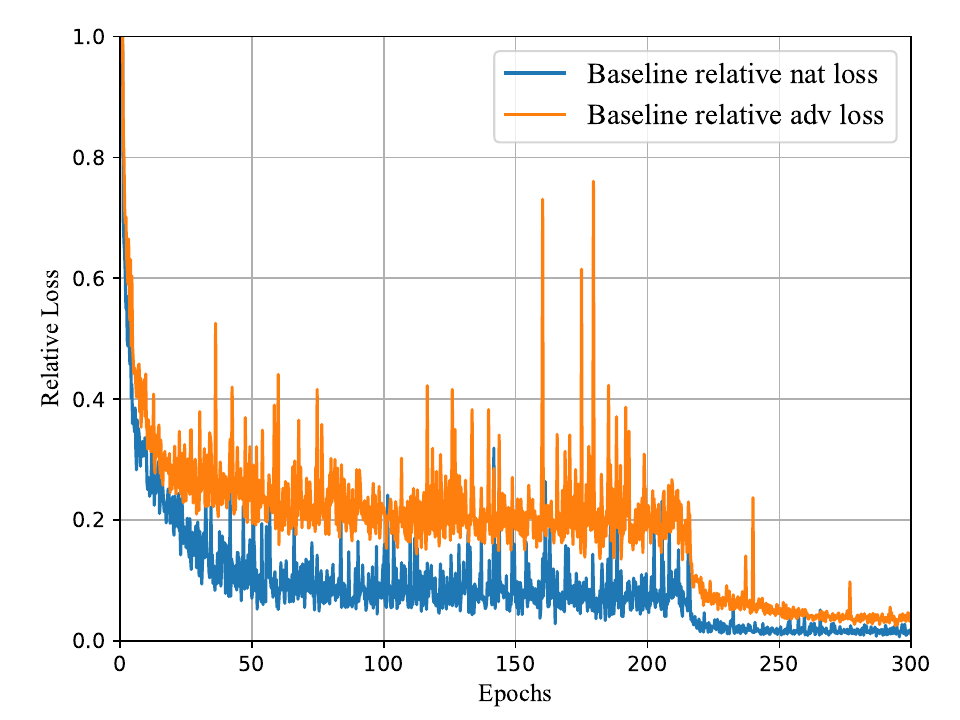}
  \includegraphics[width=0.48\linewidth]{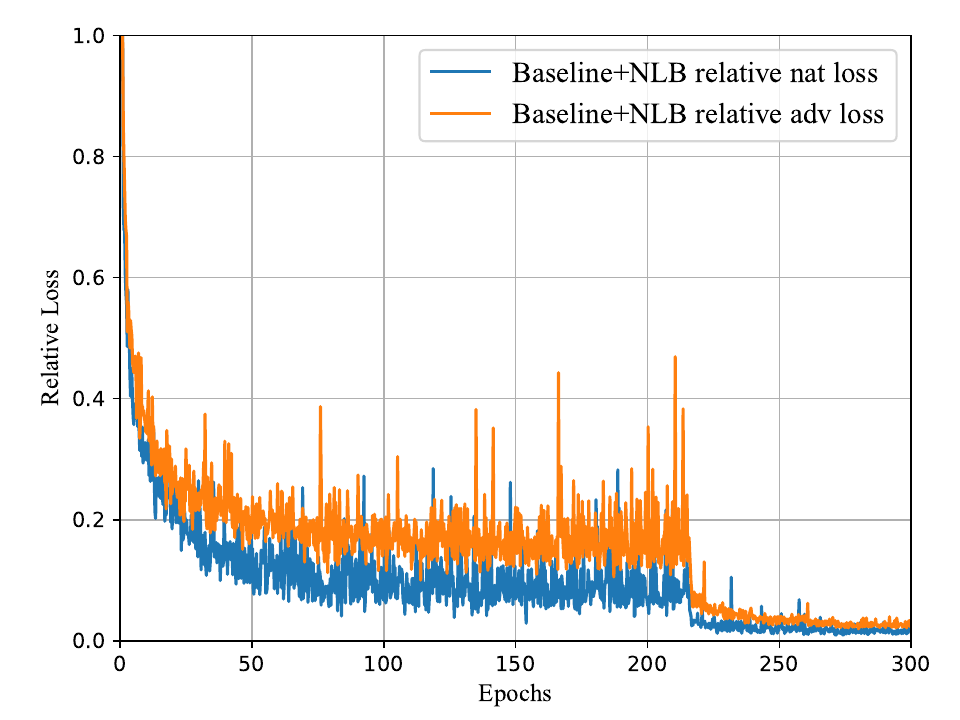}\\
\caption{The relative training loss $\tilde{L}$ with ResNet-18 student on CIFAR-10. The left is Baseline, and the right is Baseline+NLB. NLB represents Normalization Loss Balance. The x-axis means the training epochs, and the y-axis is the relative loss $\tilde{L}_{adv}$ and $\tilde{L}_{nat}$ in the training epoch x.}
\label{fig:result3}
\end{figure}

\begin{figure}[ht]
  \centering
    \includegraphics[width=0.48\linewidth]{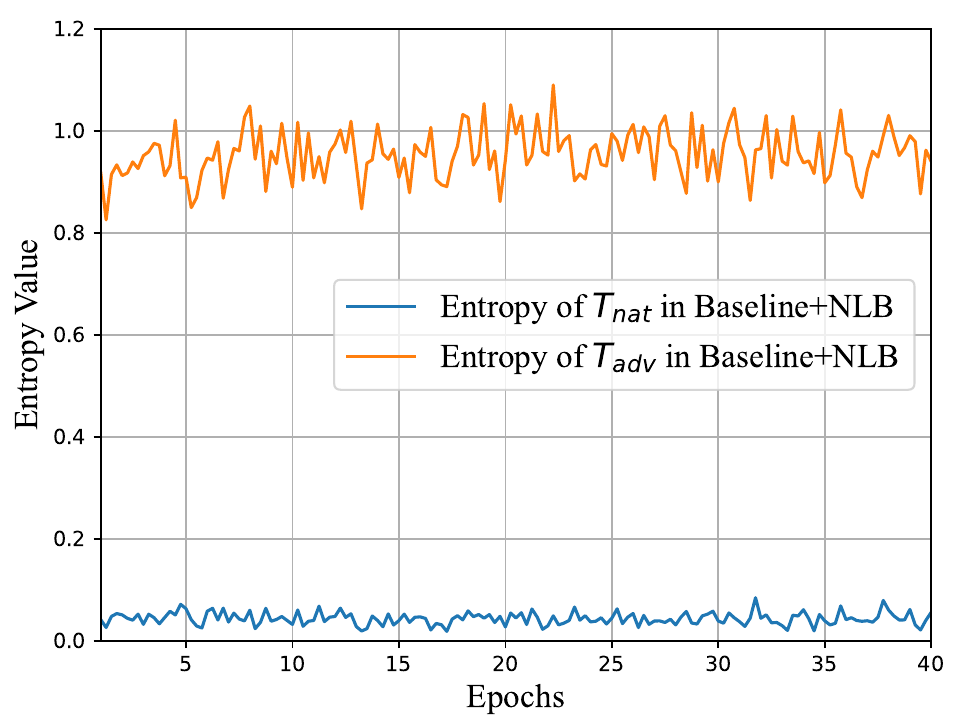}
  \includegraphics[width=0.48\linewidth]{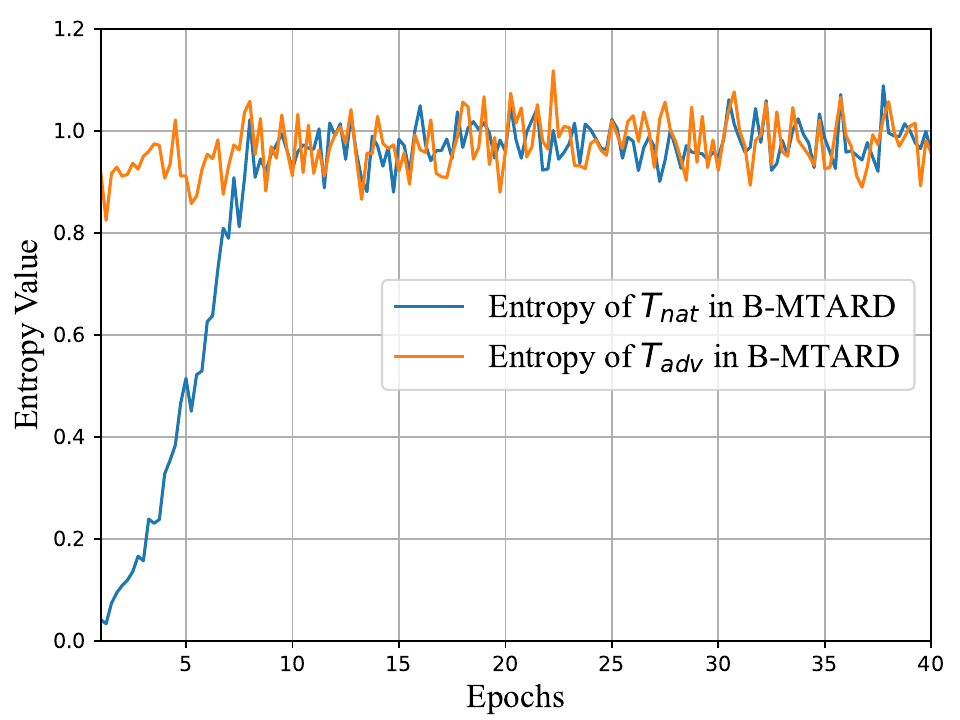}\\
\caption{The teachers' information entropy of $T_{adv}$ and $T_{nat}$ with ResNet-18 student on CIFAR-10. The left is Baseline+NLB, and the right is B-MTARD (Baseline+NLB+EBB). NLB and EBB are abbreviations of Normalization Loss Balance and Entropy-based Balance. The x-axis means the training epochs (the first 40 epochs), and the y-axis is the entropy $T_{adv}$ and $T_{nat}$ in the training epoch x.}
\label{fig:result4}
\end{figure}

To better understand the impact of each component in our B-MTARD, we conduct a set of ablation studies. 
The baseline denotes using a clean teacher and a robust teacher to guide the student from both aspects, respectively, where the weight $w_{nat}$ and $w_{adv}$ are constant at 0.5. Baseline+NLB term denotes adding the Normalization Loss Balance (NLB) algorithm to dynamically adjust the weight $w_{nat}$ and $w_{adv}$ based on the Baseline, {which is also the method in our ECCV version (MTARD) \cite{zhao2022enhanced}.} Baseline+EBB term denotes adding the Entropy-Based Balance (EBB) algorithm to dynamically adjust the $\tau_{nat}$ and $\tau_{adv}$ based on the Baseline, {where the weight $w_{nat}$ and $w_{adv}$ are constant at 0.5}. B-MTARD is the final version of our method, {which applies both Normalization Loss Balance and Entropy-Based Balance algorithm.}
The performance is shown in Fig. \ref{fig:result1}. 
The change of total loss $L_{total}$ in the training process is shown in Fig. \ref{fig:result2}, and the change of relative loss $\tilde{L}_{nat}$ and $\tilde{L}_{adv}$ in the training process is shown in Fig. \ref{fig:result3}. The change for the teacher's information entropy in the training process is shown in Fig. \ref{fig:result4}.

In Fig. \ref{fig:result1}, B-MTARD's improvement brought by each component is remarkable, which shows its effectiveness. Multiple teachers positively affect the student model to learn with both clean and robust accuracy. However, it is not enough to mitigate the trade-off between accuracy and robustness without the Entropy-based Balance algorithm and the Normalization Loss Balance algorithm, {both of them can further enhance the W-Robust performance.} 

{Baseline+NLB outperforms Baseline by 0.87\%, 0.56\%, and 0.6\% against the attack of  PGD$_{\scriptscriptstyle \rm{sat}}$,  PGD$_{\scriptscriptstyle \rm{trades}}$, and CW$_{\scriptscriptstyle \rm{\infty}}$ in the metric of W-robust Acc, while  B-MTARD outperforms Baseline+EBB by 0.33\%, 0.23\%, and 0.15\% against the attack of  PGD$_{\scriptscriptstyle \rm{sat}}$,  PGD$_{\scriptscriptstyle \rm{trades}}$, and CW$_{\scriptscriptstyle \rm{\infty}}$ in the metric of W-robust Acc.
The improvement shows the effectiveness of our Normalization Loss Balance Algorithm. }

{In addition, B-MTARD has an obvious improvement compared with our ECCV version (Baseline + NLB): Our B-MTARD improves W-Robust Acc by 0.89\%, 0.82\%, and 0.94\% compared with Baseline + NLB against the attack of PGD$_{\scriptscriptstyle \rm{sat}}$,  PGD$_{\scriptscriptstyle \rm{trades}}$, and  CW$_{\scriptscriptstyle \rm{\infty}}$. The result denotes the necessity and effectiveness of our incremental work (the Entropy-based Balance algorithm) in this version.}

In Fig. \ref{fig:result2}, compared with RSLAD and Baseline, the Baseline+NLB's training loss is less oscillating and can converge better. In Fig. \ref{fig:result3}, the gap between $\tilde{L}_{nat}$ and $\tilde{L}_{adv}$ can represent the difference of the knowledge proportion that students have not learned from the clean teacher and the robust teacher. The Baseline+NLB's trade-off between $\tilde{L}_{nat}$ and $\tilde{L}_{adv}$ are tinier. The results demonstrate Normalization Loss Balance can indeed make the student have relatively equal knowledge learning speeds from different teachers.

In Fig. \ref{fig:result4}, we can see that in Baseline+NLB, the information entropy gap between the clean teacher and the robust teacher is very obvious, but after the adjustment of our Entropy-based Balance algorithm, the information entropy gap quickly narrows and remains relatively consistent,  {and this phenomenon can denote the knowledge scale gap between clean teacher and robust teacher reduces,} which proves the effectiveness of our Entropy-based Balance algorithm. Combined with the experimental results in Fig. \ref{fig:result1}, reducing the difference in teacher's information entropy can indeed improve the performance of the student.

\subsection{The discussion about hyper-parameters}
\textbf{Hyper-parameter $\tau$}. The temperatures of different teachers are automatically adjusted by our Entropy-based algorithm until the information entropy is equal. However, the initial temperature setting still directly influences the amount of information entropy for both teachers and determines the prior adjustment scope, and a proper initial temperature can ensure temperature changes within a reasonable range to maintain the effectiveness of teachers' knowledge. 
To select the proper initial temperature value, we select initial $\tau_{nat}$ and $\tau_{adv}$  as 1, 2, 5, 8, {while all other experimental settings remain unchanged.} Based on the results shown in Table \ref{table:temp study}, we select 1 as our initial temperature value.

\begin{table}[t] 
\begin{center}
\caption{Study with different initial $\tau$. {$\beta$ is fixed to 1,} the students are ResNet-18 trained on CIFAR-10 guided by ResNet-56 and WideResNet-34-10. }
\label{table:temp study}
\begin{tabular}{m{2cm}<{\centering}|m{1.4cm}<{\centering}m{1.6cm}<{\centering}m{1.6cm}<{\centering}m{1.8cm}<{\centering}}
\hline
  Attack & initial $\tau$  & Clean & Robust& W-Robust \\
  \hline
 \multirow{4}*{PGD$_{\scriptscriptstyle \rm{sat}}$\cite{madry2017towards}} & $\tau = 1$  & 88.20\%  & 51.68\% & \textbf{69.94\%}  \\
 ~ & $\tau = 2$ & 88.22\% & 51.43\% & 69.83\% \\
  ~ & $\tau = 5$ & 89.27\% & 47.89\% & 68.58\% \\
 ~ &  $\tau = 8$ &  88.64\% & 46.97\% &  67.81\% \\
\hline
 \multirow{4}*{PGD$_{\scriptscriptstyle \rm{trades}}$\cite{zhang2019theoretically}} & $\tau = 1$  & 88.20\% & 54.40\% & \textbf{71.30\%} \\
 ~ & $\tau = 2$ & 88.22\% & 54.13\% & 71.18\% \\
  ~ & $\tau = 5$ & 89.27\% & 50.90\%  & 70.09\% \\
 ~ &  $\tau = 8$ & 88.64\% & 49.50\% & 69.07\% \\
\hline
 \multirow{4}*{CW$_{\scriptscriptstyle \rm{\infty}}$\cite{carlini2017towards}} & $\tau = 1$  & 88.20\% & 49.88\% & \textbf{69.04\%} \\
 ~ & $\tau = 2$ & 88.22\% & 49.79\% & 69.01\% \\
  ~ & $\tau = 5$ & 89.27\% & 43.99\% & 66.63\% \\
 ~ &  $\tau = 8$ & 88.64\% &  42.45\% & 65.55\% \\
\hline
\end{tabular}
\end{center}
\end{table}

\begin{table}[ht] 
\begin{center}
\caption{Study with different $\beta$. {Initial $\tau$ is fixed to 1,} the students are ResNet-18 trained on CIFAR-10 guided by ResNet-56 and WideResNet-34-10.  }
\label{table:appendix1}
\begin{tabular}{m{2cm}<{\centering}|m{1.4cm}<{\centering}m{1.6cm}<{\centering}m{1.6cm}<{\centering}m{1.8cm}<{\centering}}
\hline
  Attack & $\beta$  & Clean & Robust& W-Robust \\
  \hline
 \multirow{4}*{PGD$_{\scriptscriptstyle \rm{sat}}$\cite{madry2017towards}} & $\beta = 0.5$  & 88.51\% & 50.24\% & 69.38\% \\
 ~ & \textbf{$\beta = 1$} & 88.20\%  & 51.68\% & \textbf{69.94\%}\\
  ~ & $\beta = 4$ & 89.10\% & 50.01\% & 69.56\% \\
 ~ &  $\beta = 10$ & 88.63\% & 49.09\% & 68.86\% \\
\hline
 \multirow{4}*{PGD$_{\scriptscriptstyle \rm{trades}}$\cite{zhang2019theoretically}} & $\beta = 0.5$  & 88.51\% & 53.43\% & 70.97\%\\
 ~ & \textbf{$\beta = 1$} & 88.20\% & 54.40\% & \textbf{71.30\%} \\
  ~ & $\beta = 4$ & 89.10\% & 53.04\% & 71.07\% \\
 ~ &  $\beta = 10$ & 88.63\% & 52.63\% & 70.63\% \\
\hline
 \multirow{4}*{CW$_{\scriptscriptstyle \rm{\infty}}$\cite{carlini2017towards}} & $\beta = 0.5$  & 88.51\% & 48.59\% & 68.55\% \\
 ~ & \textbf{$\beta = 1$} & 88.20\% & 49.88\% & \textbf{69.04\%} \\
 ~ & $\beta = 4$ & 89.10\% & 48.61\% & 68.86\% \\
 ~ &  $\beta = 10$ & 88.63\% & 48.16\% & 68.40\% \\
\hline
\end{tabular}
\end{center}
\end{table}

\textbf{Hyper-parameter $\beta$}. Here, we explore the role of $\beta$ in Normalization Loss Balance (Eq. (\ref{eq:21}) and Eq. (\ref{eq:22})). 
We test the results of different values $\beta$ for the training of the student. We select $\beta$ as 0.5, 1, 4, and 10 in our experiment, {while all other experimental settings remain unchanged.} The {results are} presented in Table \ref{table:appendix1}.

From the result, $\beta$ plays an important role in Normalization Loss Balance. Choosing the right $\beta$ value can affect the student's final performance. The lower $\beta$ is suitable when the student gets a similar knowledge scale from the clean teacher and the robust teacher at the current training epoch and the student's status is not fluctuating much. The larger $\beta$ brings a more severe penalty on the student when the student places too much emphasis on one teacher's knowledge but ignores another. Based on the performance, we choose the relatively best value: $\beta = 1$ in our setting.

\subsection{Adversarial Robustness Evaluation}

\textbf{White-box Robustness.} The performances of ResNet-18 and MobileNet-v2 trained by our B-MTARD and other baseline methods under the white box attacks are shown in Table \ref{table:2} and \ref{table:3} for CIFAR-10, CIFAR-100, and Tiny-ImageNet. 

The results in Table \ref{table:2} and \ref{table:3} demonstrate that B-MTARD achieves state-of-the-art W-Robust Acc on CIFAR-10, CIFAR-100, and Tiny-ImageNet. For ResNet-18, B-MTARD improves W-Robust Acc by 1.44\%, 2.20\%, and 1.41\% compared with the best baseline method against the PGD$_{\scriptscriptstyle \rm{trades}}$ on CIFAR-10, CIFAR-100, and Tiny-ImageNet; B-MTARD improves W-Robust Acc by 0.79\%, 2.76\%, and 1.54\% compared with the best baseline method against the PGD$_{\scriptscriptstyle \rm{trades}}$ on CIFAR-10, CIFAR-100, and Tiny-ImageNet. Moreover, B-MTARD shows relevant superiority against FGSM, PGD$_{\scriptscriptstyle \rm{sat}}$, CW$_{\scriptscriptstyle \rm{\infty}}$, and AutoAttack compared with other methods. The results show that the overall performance of B-MTARD has different degrees of advantage whether compared with the other baseline methods. {Meanwhile, B-MTARD also shows the corresponding superiority compared with our conference version (MTARD).}

\textbf{Black-box Robustness.} In addition, we also test B-MTARD and other methods against black-box attacks for ResNet-18 and MobileNet-v2 on CIFAR-10, CIFAR-100, and Tiny-ImageNet separately. We choose the transfer-based attack and query-based attack in our evaluation. We select the best checkpoint of all the trained models based on W-Robust Acc. The {results} of ResNet-18 {are} shown in Table \ref{table:4}, while the {results} of MobileNet-v2 {are} shown in Table \ref{table:5}. 

From the result, B-MTARD trained models achieve the best W-Robust Acc against all three black-box attacks compared to other models. Under the Square Attack, 
B-MTARD trained ResNet-18 improves W-Robust Acc by 4.45\%, 3.58\%, and 2.74\% on CIFAR-10, CIFAR-100, and Tiny-ImageNet compared to the second-best method; Moreover, B-MTARD brings 5.56\%, 4.78\%, and 3.31\% improvements to MobileNet-v2 on CIFAR-10, CIFAR-100, and Tiny-ImageNet. In addition, B-MTARD has different margins in defending against query-based attack of RayS and transfer attacks of PGD$_{\scriptscriptstyle \rm{trades}}$ and CW$_{\scriptscriptstyle \rm{\infty}}$, which shows the superior performance of B-MTARD in black-box robustness.

\textbf{Excluding Obfuscated Gradients.}
Although some defense methods claim to resist adversarial attacks, their vulnerabilities are often exposed because they belong to  ``Obfuscated Gradients'' \cite{tramer2020adaptive,athalye2018obfuscated}.  Athalye et al. \cite{athalye2018obfuscated} have confirmed that Adversarial Training \cite{madry2017towards} can effectively defend against the adaptive attack and do not belong to ``Obfuscated Gradients''. Here we argue that B-MTARD is excluding the ``Obfuscated Gradients'' from the several aspects following \cite{athalye2018obfuscated} and \cite{wang2019improving}: (1) Our B-MTARD can effectively defend against strong attack: AutoAttack \cite{croce2020reliable}, which includes strong white-box attack: A-PGD and strong black-box attack: Square Attack \cite{andriushchenko2020square}. (2) Strong white-box test attacks (e.g., CW$_{\scriptscriptstyle \rm{\infty}}$\cite{carlini2017towards}) have higher attack success rates than weak white-box test attacks (e.g., FGSM \cite{goodfellow2014explaining}) in Table \ref{table:2} and Table \ref{table:3}. (3) White-box test attacks have higher attack success rates than Black-box test attacks (Comparing with Table \ref{table:2} and Table \ref{table:4}, Table \ref{table:3} and Table \ref{table:5}). (4) To avoid insufficient attack rounds and to fall into the local optimal solution in the general attack methods, a gradient estimation attack method from \cite{tramer2020adaptive} is directly applied to test B-MTARD. The {results} in Table \ref{table:adaptive attack} show B-MTARD can effectively resist the strong gradient estimation attack. Meanwhile, the black-box attack success rate is lower than the white-box attack, which further confirms B-MTARD is not ``Obfuscated Gradients''.
\setlength{\tabcolsep}{1pt}
\begin{table}[t] 
\begin{center}
\caption{The gradient estimation (GE) Attack \cite{tramer2020adaptive} toward the ResNet-18 trained by B-MTARD on different datasets. The results are the robust Acc.}
\label{table:adaptive attack}
\begin{tabular}{m{2.6cm}<{\centering}|m{1.6cm}<{\centering}m{1.6cm}<{\centering}m{2.2cm}<{\centering}}
\hline
  Attack & CIFAR-10 & CIFAR-100 & Tiny-ImageNet \\
\hline
 Base PGD$_{\scriptscriptstyle \rm{trades}}$\cite{zhang2019theoretically} & 54.40\% & 29.94\% & 24.94\%  \\
 GE attack \cite{tramer2020adaptive} & 60.34\% & 32.53\% & 30.83\%   \\
\hline
\end{tabular}
\end{center}
\end{table}
\setlength{\tabcolsep}{1.4pt}

\subsection{Comparison with other Multi-teacher Distillation}
{In order to further prove our effectiveness, we apply the representative multi-teacher distillation method \cite{you2017learning} into the framework of Adversarial Training to compare with our method.  Here we apply the same adversarial teacher and clean teacher with our B-MTARD for \cite{you2017learning}. We apply the weighted average of different teacher output logits as the guidance for both the adversarial examples and clean examples to train the student model (ResNet-18) on CIFAR-10. The results are shown in Table \ref{table:other multi teacher baseline}.}
\begin{table}[th] 
\begin{center}
\caption{{The Comparison with other Multi-teacher Distillation. The results are trained on CIFAR-10 of ResNet-18. The results are on W-Robust Acc.}}
\label{table:other multi teacher baseline}
\begin{tabular}{m{1.5cm}<{\centering}|m{1.2cm}<{\centering}m{1.2cm}<{\centering}m{1.3cm}<{\centering}m{1.2cm}<{\centering}m{1.2cm}<{\centering}}
\hline
  Attack & FGSM &  PGD$_{\scriptscriptstyle \rm{sat}}$ &  PGD$_{\scriptscriptstyle \rm{trades}}$ &  CW$_{\scriptscriptstyle \rm{\infty}}$&  AA \\
\hline
 LMTN\cite{you2017learning} & 72.56\% & 68.50\% & 69.62\% & 68.12\% & 67.09\%  \\
 \textbf{B-MTARD} & \textbf{74.81\%} & \textbf{69.94\%} & \textbf{71.30\%} & \textbf{69.04\%} & \textbf{67.82\%}   \\
\hline
\end{tabular}
\end{center}
\end{table}

{The results show the effectiveness of our proposed method. Our B-MTARD outperforms the LMTN by 2.25\%, 1.44\%, 1.68\%, 0.92\%, and 0.73\% against the attack of FGSM,  PGD$_{\scriptscriptstyle \rm{sat}}$,  PGD$_{\scriptscriptstyle \rm{trades}}$,  CW$_{\scriptscriptstyle \rm{\infty}}$, and AA in the metric of W-robust Acc. 
Actually, just as the discussion in related work, since the training of clean samples and adversarial samples will interfere with each other and lead to the existence of the trade-off phenomenon, how to choose an appropriate balance adjustment strategy to eliminate the existence of trade-off as much as possible is a difficult optimization challenge.
Directly applying previous multi-teacher optimization methods has shortcomings in this new task scenario, which has been proven by experimental results.
For that, we propose the Entropy-Based Balance algorithm and Normalization Loss Balance algorithm to balance the two optimization directions from the perspective of teacher and student, respectively, which do provide a balance in the conflicting multi-teacher distillation. The results further show the necessity and superiority of those two algorithms.}


\subsection{Further Explorations for Different Teacher Models}

To better understand the impact of different teachers, we evaluate the performance of the same students guided by different teachers. We choose ResNet-18 as the student, and ResNet-56 and WideResNet-34-10-C (95.83\% clean accuracy) as the clean teacher, respectively. Also, we choose WideResNet-34-10 and WideResNet-70-16 (From \cite{gowal2020uncovering} with stronger robustness) as the robust teacher, and other experimental settings are the same as the initial settings. The performance is shown in Table \ref{table:appendix7}.
 
\setlength{\tabcolsep}{1pt}
\begin{table}[t] 
\begin{center}
\caption{Study with different teacher models using our B-MTARD, WRN-34-10-C is the clean teacher model WideResNet-34-10-C,  WRN-34-10 and WRN-70-16 are the abbreviations of WideResNet-34-10 and WideResNet-70-16. RN-56 is the abbreviation of ResNet-56. All the results are the best checkpoints based on W-Robust Acc.}
\label{table:appendix7}
\begin{tabular}{m{2.0cm}<{\centering}|m{2.0cm}<{\centering}m{2.0cm}<{\centering}m{2.0cm}<{\centering}}
\hline
  Attack & Clean Teacher& Robust Teacher & W-Robust \\
  \hline
 \multirow{3}*{PGD$_{\scriptscriptstyle \rm{sat}}$\cite{madry2017towards}} &RN-56 &WRN-34-10 & \textbf{69.94\%} \\
 ~ & WRN-34-10-C &WRN-34-10  & 69.09\% \\
 ~ & WRN-34-10-C &WRN-70-16 &  68.49\%\\
\hline
 \multirow{3}*{PGD$_{\scriptscriptstyle \rm{trades}}$\cite{zhang2019theoretically}} &RN-56 &WRN-34-10 & \textbf{71.30\%} \\
 ~ & WRN-34-10-C &WRN-34-10  & 70.66\% \\
 ~ & WRN-34-10-C &WRN-70-16 &  70.01\%\\
\hline
 \multirow{3}*{CW$_{\scriptscriptstyle \rm{\infty}}$\cite{carlini2017towards}} &RN-56 &WRN-34-10 & \textbf{69.04\%} \\
 ~ & WRN-34-10-C &WRN-34-10 &  68.48\% \\
 ~ & WRN-34-10-C &WRN-70-16 & 67.89\% \\
\hline
\end{tabular}
\end{center}
\end{table}
\setlength{\tabcolsep}{1.4pt}

To our surprise, the student's ability is not improved as the teacher's parameter scale increases. Although larger teachers have stronger accuracy and robustness, the student guided by clean ResNet-56 and adversarial WideResNet-34-10 achieves the best W-Robust Acc instead of other students guided by larger teachers, which demonstrates that small teachers still have pretty potential to guide the student and can bring effective improvement to the student. Meanwhile, the performance of the student is not strictly bound to the performance of the teacher and the model size, which still has potential space to explore.

\section{Conclusion}
\label{sec:Conclusion}
This paper focused on mitigating the trade-off between accuracy and robustness in Adversarial Training. To bring both clean and robust knowledge, we proposed Balanced Multi-Teacher Adversarial Robustness Distillation (B-MTARD) to guide the student model, in which a clean teacher and a robust teacher are applied in a balanced state during the Adversarial Training process. In addition, we proposed the Entropy-Based Balance algorithm to keep the knowledge scale of these teachers consistent. In order to ensure that the student gets equal knowledge from two teachers, we designed a method to use Normalization Loss Balance in B-MTARD. A series of experiments demonstrated that our B-MTARD outperformed the existing Adversarial Training methods and adversarial robustness distillation on CIFAR-10, CIFAR-100, and Tiny-ImageNet. 
In the future, B-MTARD can be applied to other tasks with multiple optimization goals, not only limited to adversarial training, which has great development potential.

\section*{Acknowledgement}
This work was supported by the Project of the National Natural Science Foundation of China (No.62076018), and the Fundamental Research Funds for the Central Universities.

\ifCLASSOPTIONcaptionsoff
  \newpage
\fi

\bibliographystyle{splncs04}
\bibliography{egbib}

\vspace{-1cm}
\begin{IEEEbiography}[{\includegraphics[width=1in,height=1.25in,clip,keepaspectratio]{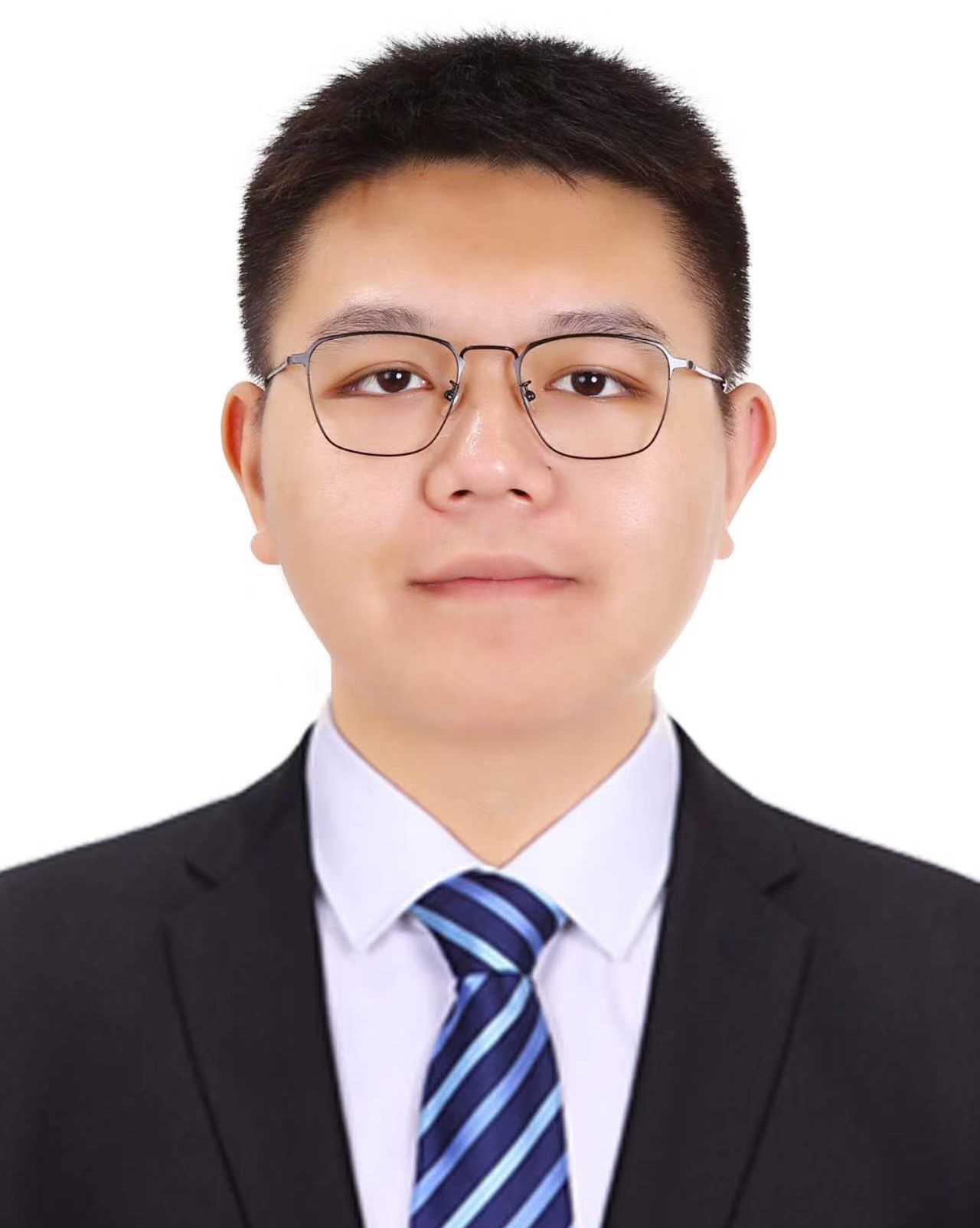}}]{Shiji Zhao} received his B.S. degree in the School of Computer Science and Engineering, Beihang University (BUAA), China. He is now a Ph.D student in the Institute of Artificial Intelligence, Beihang University (BUAA), China. His research interests include computer vision, deep learning and adversarial robustness in machine learning.
\end{IEEEbiography}
\vspace{-1cm}
\begin{IEEEbiography}[{\includegraphics[width=1in,height=1.25in,clip,keepaspectratio]{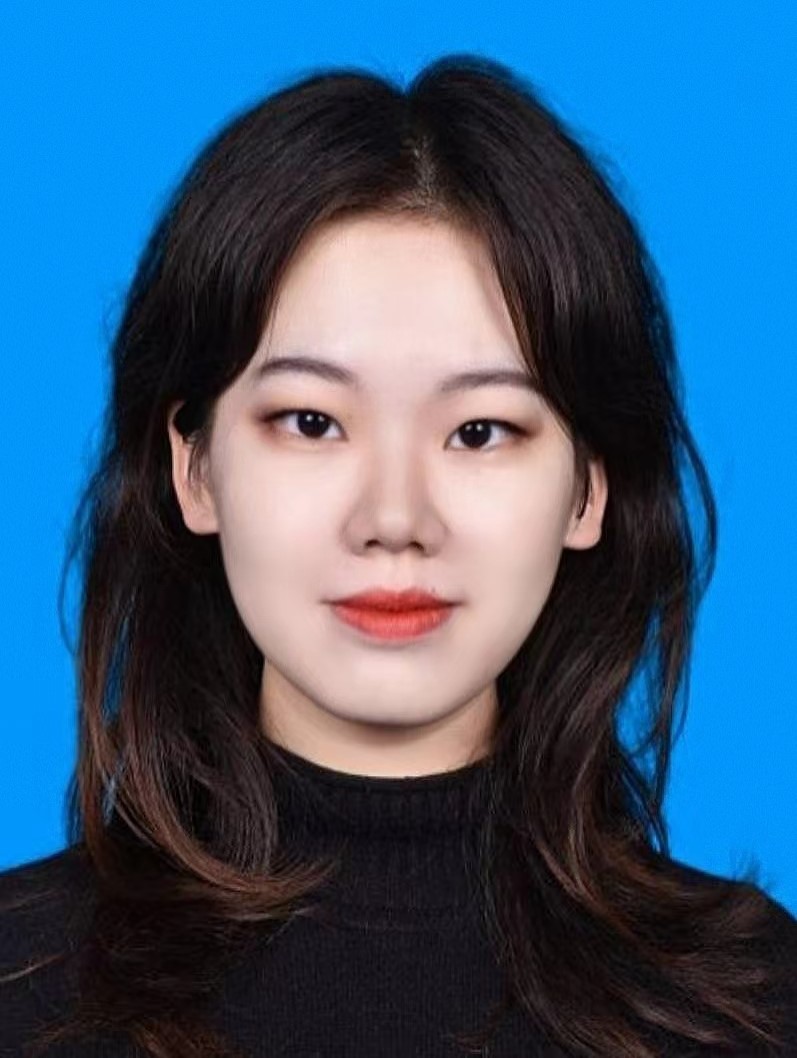}}]{Xizhe Wang} received her B.S. degree in the Institute of Artificial Intelligence, Beihang University (BUAA), China. She is now an M.S. student in the Institute of Artificial Intelligence, Beihang University (BUAA), China. Her research interests include computer vision, deep learning and adversarial robustness in machine learning.
\end{IEEEbiography}
\vspace{-1cm}
\begin{IEEEbiography}[{\includegraphics[width=1in,height=1.25in,clip,keepaspectratio]{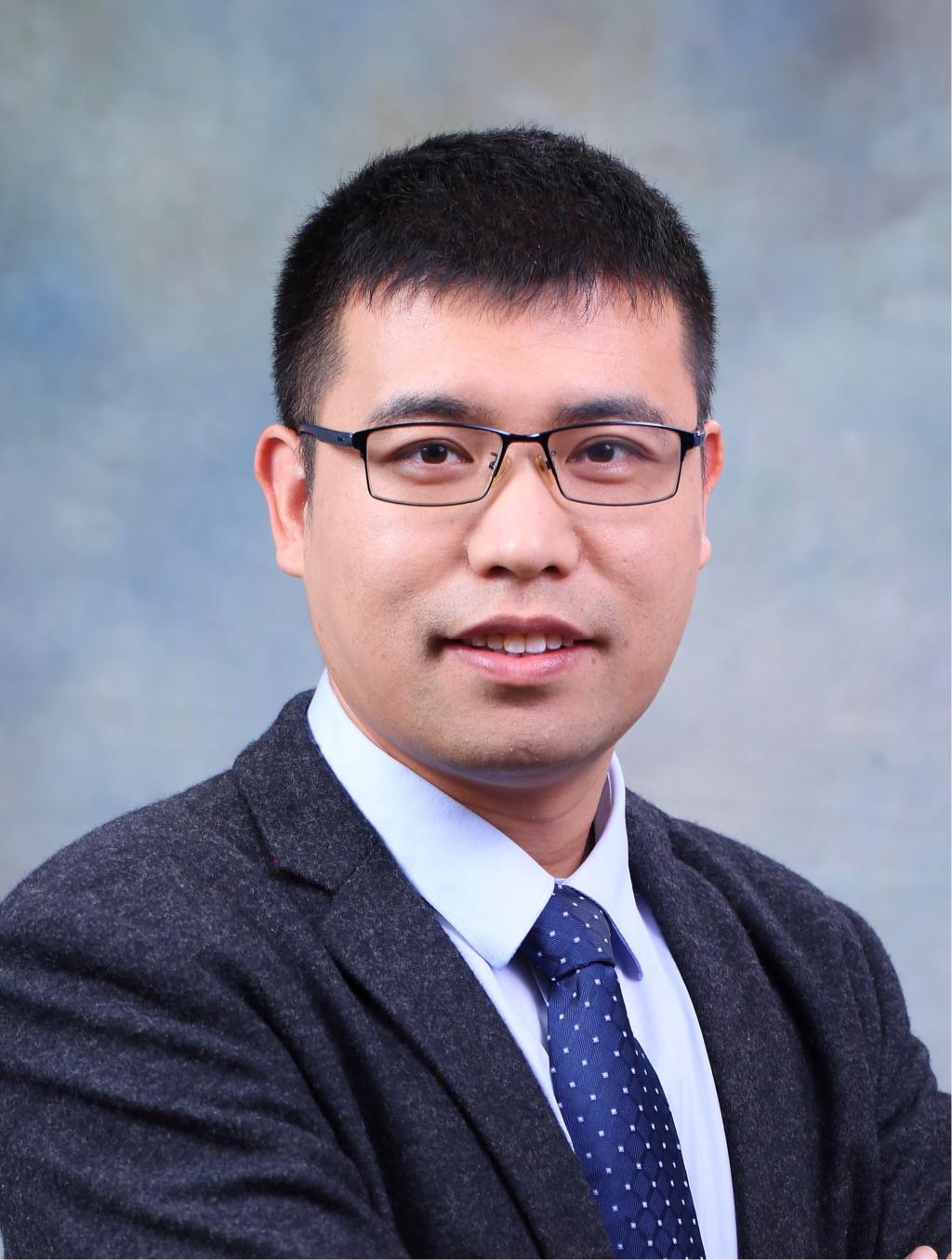}}]{Xingxing Wei} received his Ph.D degree in computer science from Tianjin University, and B.S.degree in Automation from Beihang University (BUAA), China. He is now an Associate Professor at Beihang University (BUAA). His research interests include computer vision, adversarial machine learning and its applications to multimedia content analysis. He is the author of referred
journals and conferences in IEEE TPAMI,  IJCV, CVPR, ICCV, ECCV,  etc.
\end{IEEEbiography}

\title{Mitigating Accuracy-Robustness Trade-off via Balanced Multi-Teacher Adversarial Distillation: Appendix}

\maketitle

\IEEEdisplaynontitleabstractindextext

%

\appendices
\section{The Proof}
\begin{proof}
\label{proof}
\emph{
Based on the definition of relation entropy, the knowledge scale $K^{T}$ can further expend as follows:
\begin{align}
\label{knowledge defination}
K^{T}=KL(I(x),T(x))= -\sum_{k=1}^Cp_k^{T}(x)log(p_k^{I}(x))-H(P^{T}), 
\end{align}
where $P^{I}=\{p_{1}^{I}(x),..., p_{C}^{I}(x)\}$ and $P^{T}=\{p_{1}^{T}(x),..., p_{C}^{T}(x)\}$ denote the general model $I$ predicted distribution and the well-trained teacher model $T$ predicted distribution, respectively. 
For the general model, the model's prediction has no preference towards any samples, so a reasonable assumption can be made as follows:
\begin{align}
\label{eq:7}
p_{k}^{I}(x)=1/C, k = 1,2,..., C.
\end{align}
Based on this assumption, we can easily find that the first term ($ -\sum_{k=1}^Cp_k^{T}(x)log(p_k^{I}(x))$) in Eq.(\ref{knowledge defination}) satisfies:
\begin{align}
-\sum_{k=1}^Cp_k^{T}(x)log(p_k^{I}(x)) = -\sum_{k=1}^Cp_k^{T}(x)log(1/C),
\end{align}
because the sum of the model's prediction is equal to 1, so $\sum_{k=1}^Cp_k^{T}(x) = 1$, then we have:
\begin{align}
-\sum_{k=1}^Cp_k^{T}(x)log(p_k^{I}(x)) = -log(1/C)=logC,
\end{align}
and $-log(1/C)$ is a constant that does not change, so the teacher’s knowledge scale is negatively related to the second term in Eq.(\ref{knowledge defination}) (information entropy $H(P^{T})$) as follows:
\begin{align}
\label{knowledge with entropy}
K^{T}=logC-H(P^{T}). 
\end{align}
Then the Theorem 1 is proved.
}
\end{proof}

\label{proofA}
\begin{proof}
\emph{
Here we firstly define the $q$ as $exp(z(x)/\tau)$ in Eq. (3) of main text, and $p_m$= $q_m/\sum_{j=1}^C q_j$, and the distribution $P=\{p_{1}, p_{2},..., p_{C}\}$ , then we further expand information entropy $H(P)$ as follows:
\begin{align}
    &H(P)=-\sum_{j=1}^Cp_jlog(p_j), \\
                &= -\sum_{j=1}^C(\frac{q_j}{\sum_{k=1}^C q_k})log(\frac{q_j}{\sum_{k=1}^C q_k}), \\
                &=-\frac{[(\sum_{j=1}^C q_j log q_j)-(\sum_{j=1}^Cq_j)log(\sum_{j=1}^Cq_j)]}{ \sum_{j=1}^Cq_j}, \\
                &=-\frac{(\sum_{j=1}^C q_j log q_j)} {\sum_{j=1}^C q_j} + log \sum_{j=1}^C q_j.
\end{align}
After the information entropy $H(P)$ is decomposed, we first try to solve the partial derivative of information entropy $H(P)$ with respect to $q_m$ as follows:
\begin{align}
    &\nabla_{q_m}H(P) \notag \\
    &=\frac{1}{ \sum_{j=1}^C q_j}-\frac{((log q_m +1)\sum_{j=1}^C q_j - \sum_{j=1}^C q_j log q_j )}{ (\sum_{j=1}^C q_j)^2},   \\
    &= -\frac{(log q_m \sum_{j=1}^C q_j - \sum_{j=1}^C q_j log q_j)} {(\sum_{j=1}^C q_j)^2}, \\
    &= - \frac{log q_m}{\sum_{j=1}^C q_j} + \frac{\sum_{j=1}^C q_j log q_j}{(\sum_{j=1}^C q_j)^2}.
\end{align}
Then we compute the partial derivative of $q_m$ with respect to $\tau$ as follows:
\begin{align}
    \nabla_{\tau}q_m    = - \frac{q_m z_m(x)}{\tau^2}
                        = - \frac{q_m log q_m }{\tau}.
\end{align}
According to the chain rule of derivation, we can compute the partial derivative of information entropy with respect to $\tau$ as follows:
\begin{align}
    &\nabla_{\tau}H(P)    =\sum_{m=1}^C \nabla_{q_m}H(P) \nabla_{\tau}q_m\\  
    &= \sum_{m=1}^C[\frac{- log q_m}{\sum_{j=1}^C q_j} + \frac{\sum_{j=1}^C q_j log q_j}{ (\sum_{j=1}^C q_j)^2}] (- \frac{q_m log q_m }{\tau})\\
    &= \frac{1}{\tau}\sum_{m=1}^C[\frac{q_mlog^2q_m}{ \sum_{j=1}^C q_j} - \frac{q_m log q_m \sum_{j=1}^C q_j log q_j }{ (\sum_{j=1}^C q_j)^2}] \\
    &=\frac{[(\sum_{j=1}^Cq_{j})(\sum_{j=1}^Cq_{j}log^2q_{j})-(\sum_{j=1}^Cq_{j}logq_{j})^2]}{\tau(\sum_{j=1}^Cq_{j})^2}
\end{align}
}
\end{proof}

\begin{proof}
\emph{
\label{proofB}
Here we try to prove the $\nabla_{\tau}H(P) \geq 0$, according to Cauchy-Buniakowsky-Schwarz inequality, the derivation is as follows:
\begin{align}
&(\sum_{j=1}^Cq_{j})(\sum_{j=1}^Cq_{j}log^2q_{j}), \\
&=(\sum_{j=1}^C(\sqrt{q_{j}})^2)(\sum_{j=1}^C((\sqrt{q_{j}})logq_{j})^2), \\
&\geq (\sum_{j=1}^Cq_{j}logq_{j})^2.
\end{align}
The equal sign $=$ holds only if all $q_j$ are equal, and the $\tau$ and $q_i$ is always more than 0, 
\begin{align}
&\nabla_{\tau}H(P)  \notag \\
&=\frac{(\sum_{j=1}^Cq_{j})(\sum_{j=1}^Cq_{j}log^2q_{j})-(\sum_{j=1}^Cq_{j}logq_{j})^2}{\tau(\sum_{j=1}^Cq_{j})^2} 
\geq 0.
\end{align}
Then conclusion $\nabla_{\tau}H(P) \geq 0$ is proved.
}
\end{proof}

\ifCLASSOPTIONcaptionsoff
  \newpage
\fi
\end{document}